%% file: main.tex
\documentclass[runningheads]{llncs}

\usepackage{eccv}

\usepackage{eccvabbrv}

\usepackage{graphicx}
\usepackage{booktabs}

\usepackage[accsupp]{axessibility}  %

\input{src/00.Preamble}

\usepackage[pagebackref,breaklinks,colorlinks,citecolor=eccvblue]{hyperref}

\usepackage{orcidlink}

\begin{document}

\title{This Probably Looks \emph{Exactly} Like That:\\An Invertible Prototypical Network}

\titlerunning{This Probably Looks \emph{Exactly} Like That}

\author{Zachariah Carmichael*\orcidlink{0000-0002-7603-2004} \and
Timothy Redgrave*\orcidlink{0000-0003-2425-8543} \and
Daniel Gonzalez Cedre*\orcidlink{0000-0002-2676-1452} \and
Walter J. Scheirer\orcidlink{0000-0001-9649-8074}}

\authorrunning{Z.~Carmichael et al.}

\institute{University of Notre Dame, Notre Dame IN 46556, USA\\
\email{zcarmich@alumni.nd.edu,\{tredgrav,dgonza26,wscheire\}@nd.edu}\\
\url{https://github.com/craymichael/ProtoFlow}
\\
*\textsc{equal contribution}
}

\maketitle

\begin{abstract}

    We combine concept-based neural networks with generative, flow-based classifiers into a novel, intrinsically explainable, exactly invertible approach to supervised learning. Prototypical neural networks, a type of concept-based neural network, represent an exciting way forward in realizing human-comprehensible machine learning without concept annotations, but a human-machine semantic gap continues to haunt current approaches. We find that reliance on indirect interpretation functions for prototypical explanations imposes a severe limit on prototypes' informative power. From this, we posit that invertibly learning prototypes as \emph{distributions} over the latent space provides more robust, expressive, and interpretable modeling. We propose one such model, called \Ours{}, by composing a normalizing flow with Gaussian mixture models. \Ours{} (1) sets a new state-of-the-art in joint generative and predictive modeling and (2) achieves predictive performance comparable to existing prototypical neural networks while enabling richer interpretation.
  \keywords{Normalizing flow \and Prototypical neural networks \and XAI}
\end{abstract}

\input{src/01.Introduction}
\input{src/02.Background}
\input{src/03.Methodology}
\input{src/04.Experiments}
\input{src/05.Discussion}

\section*{Acknowledgements}
Timothy Redgrave's travel was supported by the U.S.\ Department of Defense (Contract No.~W52P1J-20-9-3009).
The views and conclusions contained in this document are those of the authors and should not be interpreted as representing the official policies, either expressed or implied, of the U.S.\ Department of Defense or the U.S.\ Government.
The U.S.\ Government is authorized to reproduce and distribute reprints for Government purposes, notwithstanding any copyright notation here on.

\bibliographystyle{splncs04}
\bibliography{main}

\clearpage
\input{src/99.Supplemental}

\end{document}

%% file: src/00.Preamble.tex
\usepackage{xfrac}
\usepackage{bm}
\usepackage{multirow}
\usepackage{enumitem}
\usepackage{mathtools}
\usepackage{makecell}
\usepackage{stmaryrd}
\usepackage{fontawesome}
\usepackage{tensor}
\usepackage{comment}
\usepackage{pifont}%

\usepackage{pgfplots}
\usetikzlibrary{pgfplots.groupplots}
\usetikzlibrary{pgfplots.statistics}
\usepgfplotslibrary{groupplots}
\usepgfplotslibrary{fillbetween}

\usepackage{tikz}
\usepgflibrary{shadings}
\usetikzlibrary{positioning}
\usetikzlibrary{arrows}
\usetikzlibrary{arrows.meta}
\usetikzlibrary{calc}
\usetikzlibrary{angles}
\usetikzlibrary{backgrounds,fit,positioning}
\usetikzlibrary{calc}
\usetikzlibrary{math}
\usetikzlibrary{hobby}
\usetikzlibrary{patterns}
\usetikzlibrary{shapes}
\usetikzlibrary{decorations.markings,decorations.pathmorphing}
\usetikzlibrary{shapes.arrows,shapes.geometric}

\tikzstyle{wavy} = [decorate,
                    decoration={snake, amplitude=1mm, segment length=6.1mm, post length=1mm, pre length=0.0mm}]

\newcommand{\distro}[5][40]{
  \begin{tikzpicture}[line cap=round]
    \draw[#5, line width=0.1cm, name path=dense] plot[variable=\t, domain=-1:1, samples=#1, opacity=0] ({\t}, {#2 * exp(-10*(\t)^2) + #3 * exp(-60*(\t-0.6)^2 - \t) + #3 * exp(-60*(\t+0.7)^2 - 0.2) + #4 * 0.5 * exp(-50*(\t+0.3)^2) + #4 * exp(-50*(\t-0.2)^2 + 0.1)});
    \draw[-, very thick, solid, name path=xaxis] (-1, 0)--(1, 0);
    \scoped[on background layer]{
      \draw[-, very thick, solid, name path=yaxis] (0, -0.25)--(0, 1.25);
      \tikzfillbetween[of=dense and xaxis]{#5!66!background};
    }
    \draw[#5, line width=0.1cm] plot[variable=\t, domain=-1:1, samples=#1] ({\t}, {#2 * exp(-10*(\t)^2) + #3 * exp(-60*(\t-0.6)^2 - \t) + #3 * exp(-60*(\t+0.7)^2 - 0.2) + #4 * 0.5 * exp(-50*(\t+0.3)^2) + #4 * exp(-50*(\t-0.2)^2 + 0.1)});
  \end{tikzpicture}
}

\newcommand\gauss[3]{1/(#3*sqrt(2*pi))*exp(-((#1-#2)^2)/(2*#3^2))}
\newcommand\gaussian[8]{
  \begin{tikzpicture}[line cap=round]
    \draw[line width=0.1cm, draw=#8, name path=dense] plot[smooth, domain=-#1:#1, variable=\x] (\x, {0.333*\gauss{\x}{#2}{#3} + 0.334\gauss{\x}{#4}{#5} + 0.333\gauss{\x}{#6}{#7}});
    \scoped[opacity=1, transparency group, on background layer]{
      \draw[solid, very thick, draw=#8!66!background, name path=xaxis] (-#1, 0)--(#1, 0);
      \tikzfillbetween[of=dense and xaxis]{#8!66!background};
    }
  \end{tikzpicture}
}

\tikzset{shifted path/.style args={from #1 to #2 by #3}{insert path={
let \p1=($(#1.east)-(#1.center)$),
\p2=($(#2.east)-(#2.center)$),\p3=($(#1.center)-(#2.center)$),
\n1={veclen(\x1,\y1)},\n2={veclen(\x2,\y2)},\n3={atan2(\y3,\x3)} in
(#1.{\n3+180+asin(#3/\n1)}) to (#2.{\n3-asin(#3/\n2)})
}}}

\DeclareMathOperator*{\argmax}{\arg\!\max}

\newcommand{\Ours}{{ProtoFlow}}

\newcommand{\ra}[1]{\renewcommand{\arraystretch}{#1}}

\newcommand{\mat}[1]{\bm{#1}}

\DeclareMathAlphabet{\mathsfit}{\encodingdefault}{\sfdefault}{m}{sl}
\SetMathAlphabet{\mathsfit}{bold}{\encodingdefault}{\sfdefault}{bx}{n}

\newcommand{\random}[1]{\textbf{#1}}

\def\loss{{\mathcal{L}}}

\def\score{{\mathcal{S}}}
\def\proto{{\mathcal{G}}}
\def\weight{{\mat\pi}}
\def\normal{{\mathcal{N}}}
\def\softmax{{\sigma}}

\def\embedding{{\mat{z}}}

\def\data{{\mat{x}}}
\def\noise{{\random{e}}}
\def\gen{{\mat{\tilde{x}}}}
\def\class{{c}}
\def\pred{{\hat{c}}}
\def\covariance{{\mat{\Sigma}}}
\def\mean{{\mat{\mu}}}
\def\Xdomain{{\mathcal{X}}}

\def\model{{f}}

\def\trans{{\mathsf{T}}}

\def\ce{{\textsc{ce}}}
\def\elbo{{\textsc{elbo}}}
\def\con{{\textsc{cr}}}
\def\div{{\textsc{div}}}
\def\rob{{\textsc{rob}}}
\def\hell{{\mathcal{H}^2}}
\def\hellmod{{\mathcal{\widetilde{H}}^2}}

\def\domain{{\mathcal{X}}}
\def\latent{{\mathcal{Z}}}
\def\classes{{\mathcal{C}}}
\def\components{{\mathcal{K}}}

\def\platent{{p_\latent}}
\def\pdata{{p_\domain}}

\newcommand\wcon[1]{{#1}^{\bullet}}
\newcommand\wocon[1]{{#1}^{\circ}}

\newcommand{\xto}[1]{\xrightarrow{#1}}
\newcommand{\xfrom}[1]{\xleftarrow{#1}}

\newcommand\inv[1]{#1^{\scalebox{0.5}{$-1$}}}

\newcommand\ppart[5]{\tensor*[_#1^{#3}]{#5}{_{#2}^{#4}}}

\colorlet{background}{white}
\colorlet{foreground}{black}

\definecolor{stone}{HTML}{294d4a}
\definecolor{pumice}{HTML}{622227}  %
\definecolor{brown}{HTML}{782f40}

\definecolor{plotblue}{HTML}{c52e65}
\definecolor{plotred}{HTML}{3287d1}

\definecolor{blue}{HTML}{1e88e5}
\definecolor{green}{HTML}{22ac7b}  %
\definecolor{yellow}{HTML}{ffc107}
\definecolor{orange}{HTML}{ff5430}  %
\definecolor{red}{HTML}{d81b60}
\definecolor{magenta}{HTML}{d81b60}
\definecolor{purple}{HTML}{7c1d6f}

\newcommand{\todo}[1]{}

\newcommand{\suchthat}{\;\mid\;}

\DeclarePairedDelimiterX \set[1]{\lbrace}{\rbrace}{%
  \renewcommand{\suchthat}{\;\delimsize\vert\nonscript\;\mathopen{}}%
  #1%
}

\newcommand{\ML}{\textit{ML}}
\newcommand{\E}{\mathbb{E}}
\newcommand{\expval}[1]{\mathbb{E}\sq*{#1}}

\DeclarePairedDelimiter \sq{[}{]}

\DeclarePairedDelimiterX \inner[2]{\langle}{\rangle}{#1,#2}

\DeclarePairedDelimiter \norm{\lVert}{\rVert}

\usepackage{bbm, dsfont}
\AtBeginDocument{%
  \DeclareSymbolFont{AMSb}{U}{msb}{m}{n}
  \DeclareSymbolFontAlphabet{\mathbb}{AMSb}
  \DeclareMathAlphabet{\mathbb}{U}{bbold}{m}{n}
}

\makeatletter
\renewcommand\paragraph{\@startsection{paragraph}{4}{\z@}%
                       {-5\p@ \@plus -2\p@ \@minus -2\p@}%
                       {-0.75em \@plus -0.22em \@minus -0.1em}%
                       {\normalfont\normalsize\itshape}}
\makeatother

\newcommand{\cmark}{\ding{51}}
\newcommand{\xmark}{\ding{55}}

\tikzstyle{node} = [very thick,
                    circle,
                    draw=black,
                    fill=black,
                    text opacity=1]

%% file: src/01.Introduction.tex
\section{Introduction}
\label{sec:intro}

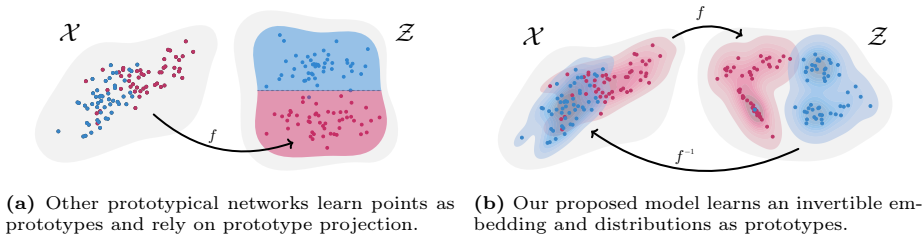
\begin{figure}[t]
  \centering
  \begin{subfigure}{.49\linewidth}
    \centering
    \resizebox{5.75cm}{!}{\input{tikz/fig1l}}
    \caption{Other prototypical networks learn points as prototypes and rely on prototype projection.}
    \label{fig:proto-points}
  \end{subfigure}
  \hfill
  \begin{subfigure}{.49\linewidth}
    \centering
    \resizebox{5.9cm}{!}{\input{tikz/fig1r}}
    \caption{Our proposed model learns an invertible embedding and distributions as prototypes.}
    \label{fig:proto-dists}
  \end{subfigure}
  \caption{%
    Existing prototypical networks rely on prototypical points and limiting means of visualizing prototypes.
    Our proposed approach, \Ours{}, enables the learning of prototypical distributions with inverses, enabling their exact and efficient visualization---\emph{\Ours{} inherently enables richer prototype interpretation}.
  }
  \label{fig:teaser}
\end{figure}

Concept-based neural networks offer an attractive way of parsing decisions made by complex systems.
By providing explanations in terms of higher-level abstractions, these models provide semantic clarity to both experts and non-experts.
While there are myriad different interpretations of the word \emph{concept} in the literature~\cite{poeta2023concept}, we are interested particularly in prototypical concepts.
Prototypes aim to distill traits not directly scrutinizable from the raw data.
For instance, the prediction of an image as a bird could be explained by its similarity to a beak prototype~\cite{protopnet}.
Explainability-by-design guarantees faithful explanations by requiring prototype involvement in decision-making~\cite{DBLP:conf/xai/PocheHB23}.
Such explanations naturally result from the symbolic form of the model~\cite{razIMLForm2022} rather than \textit{post hoc} correlational analyses~\cite{falsifiable2020}, a preferable situation for most users~\cite{kim2022hive,davoodi2023interpretability,dawoud2023human}.

Existing prototypical networks have demonstrated a human-machine semantic similarity gap and often learn irrelevant prototypes~\cite{DBLP:conf/visapp/SinhamahapatraH23,kim2022hive,protoAttack}.
These networks typically learn prototypes as \emph{points} in latent space, with limited interpretive power in data space~\cite{DBLP:conf/xai/PocheHB23,Carmichael_2024_WACV,kim2022hive,sacha2023interpretability}.
We hypothesize that this focus on \emph{points} rather than \emph{distributions} is an underlying issue; given a prototype of a blue jay, could we say the model is representing its blue color, its distinctive shape, the texture of its feathers, or something entirely different?
It's impossible to tell from just one point.
We propose to instead learn prototypical \emph{distributions} over latent space with normalizing flows, capable of generating data while providing exact likelihoods~\cite{rezende2015variational,papamakarios2021normalizing}.
They are also \emph{fully invertible,} furnishing the latent space with a faithful interpretation back to the data.
We leverage this inverse transformation to provide meaningful insight into exactly how the model is learning to represent its training data, in turn allowing an understanding of latent prototypical distributions through their samples' corresponding data-space interpretations.
This also removes constraints on learned prototypes~\cite{protopnet,DBLP:conf/xai/PocheHB23} and limitations on prototype visualizations~\cite{li2018deep,ma2024looks}.
See \cref{fig:teaser} for an overview.

In this paper, we propose a novel approach to supervised learning that is interpretable by design by bridging together key ideas from the explainable AI and generative AI literature: generative classifiers and concept-based neural networks.
We make the following contributions:
\begin{itemize}[label=\textbullet]
  \item
    We develop an approach to learning interpretable latent prototypical distributions with a joint generative and predictive model.
  \item
    We propose a diversity loss to reduce prototypical distribution overlap.
  \item
    We evaluate our method on various image classification datasets and show state-of-the-art performance at joint predictive and generative modeling.
  \item
    We qualitatively and quantitatively analyze explanations generated by the model.
    We obtain predictive performance comparable to existing prototypical neural networks while enabling richer interpretation.
\end{itemize}

%% file: tikz/fig1l.tex
\begin{tikzpicture}[ultra thick, line cap=round]
  \node at (-13, 0) {\phantom{\strut henlo}};
  \node at (13, 0) {\phantom{\strut henlo}};
  \begin{scope}
      \clip[use Hobby shortcut, closed=true]
        ( 5 - 6,  2) ..
        ( 5.25 - 6,  3) ..
        ( 0 - 6,  3) ..
        (-2 - 6,  3) ..
        (-5 - 6, -0) ..
        (-6 - 6, -3.5) ..
        ( 0 - 6, -3.25) ..
        ( 1 - 6, -3) ..
        ( 3 - 6, -2.5);
        
      \node at (-6, 0) (x1) {\includegraphics[]{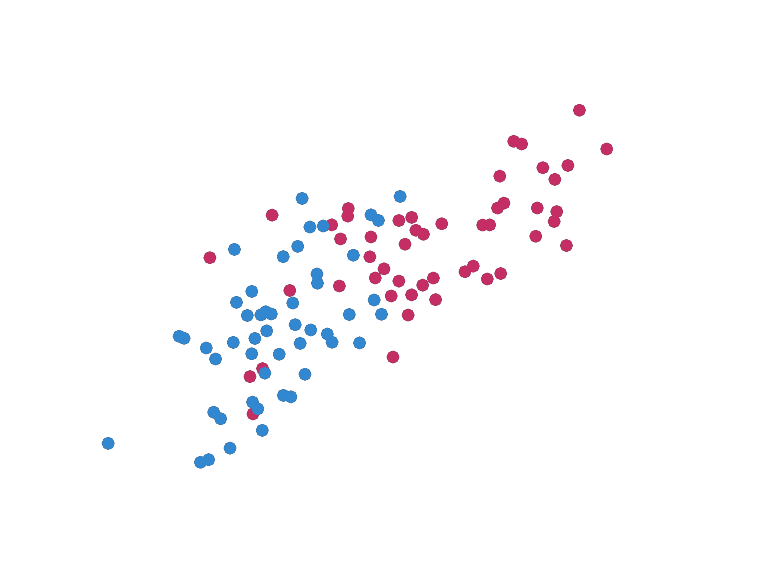}};
      \fill[gray, opacity=0.1] (current bounding box.south west) rectangle (current bounding box.north east);      
  \end{scope}

  \begin{scope}
      \clip[use Hobby shortcut, closed=true]
        (6 + 3.5, 0) ..
        (6 + 3.425, 1) ..
        (6 + 2.775, 3.25) ..
        (6 + -0.5, 3.65) ..
        (6 + -4, 3.65) ..
        (6 + -4.25, -0) ..
        (6 + -4.1, -3.25) ..
        (6 + 0, -4.1) ..
        (6 + 1, -4.1) ..
        (6 + 2.85, -3.75) ..
        (6 + 3.85, -2);

      \node at (1*6, 0) (z1) {\includegraphics[]{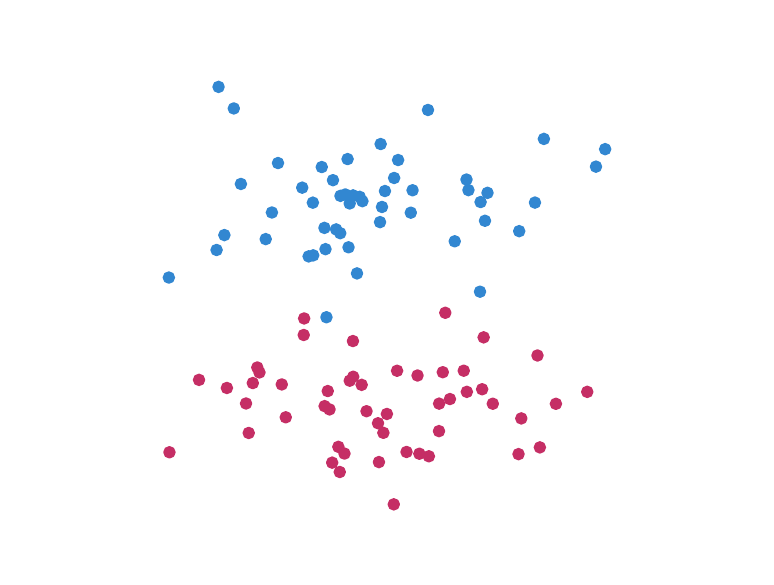}};
      \fill[red, opacity=0.45] (current bounding box.south east) rectangle (0, 0);
      \fill[blue, opacity=0.45] (0, 0) rectangle (current bounding box.north east);

      \draw[-, line width=0.075cm, loosely dashed, opacity=0.25] (0.1, 0) to (11.9, 0);
  \end{scope}
  \begin{scope}
      \clip[use Hobby shortcut, closed=true]
        (6 + 3.9, 0) ..
        (6 + 3.65, 3.25) ..
        (6 + 3.25, 3.85) ..
        (6 + 0, 4.5) ..
        (6 + -5.25, 4.25) ..
        (6 + -5.333, 0) ..
        (6 + -4.75, -4) ..
        (6 + 2, -4.5) ..
        (6 + 3, -4.5) ..
        (6 + 4, -4) ..
        (6 + 4.25, -1);
      \fill[gray, opacity=0.1] (current bounding box.south west) rectangle (current bounding box.north east);      
  \end{scope}
  
  \draw[->, bend right, line width=0.125cm] (-4.75, -1.5) to node[above, midway, xshift=-12, yshift=6]{\scalebox{3}{$f$}} (4, -3.25);
  
  \node[inner sep=0, outer sep=0] at (-10, 3.5) {\scalebox{5}{\strut $\domain$}};
  \node[inner sep=0, outer sep=0] at (11, 3.5) {\scalebox{5}{\strut $\latent$}};
    
\end{tikzpicture}

%% file: tikz/fig1r.tex
\begin{tikzpicture}[ultra thick, line cap=round]
  \node at (-13, 0) {\phantom{\strut henlo}};
  \node at (13, 0) {\phantom{\strut henlo}};
  \begin{scope}
      \clip[use Hobby shortcut, closed=true]
        ( 5 - 6,  2) ..
        ( 5.25 - 6,  3) ..
        ( 0 - 6,  3) ..
        (-2 - 6,  3) ..
        (-5 - 6, -0) ..
        (-6 - 6, -3.5) ..
        ( 0 - 6, -3.25) ..
        ( 1 - 6, -3) ..
        ( 3 - 6, -2.5);
        
      \node at ( -1*6,  0) (x2) {\includegraphics[]{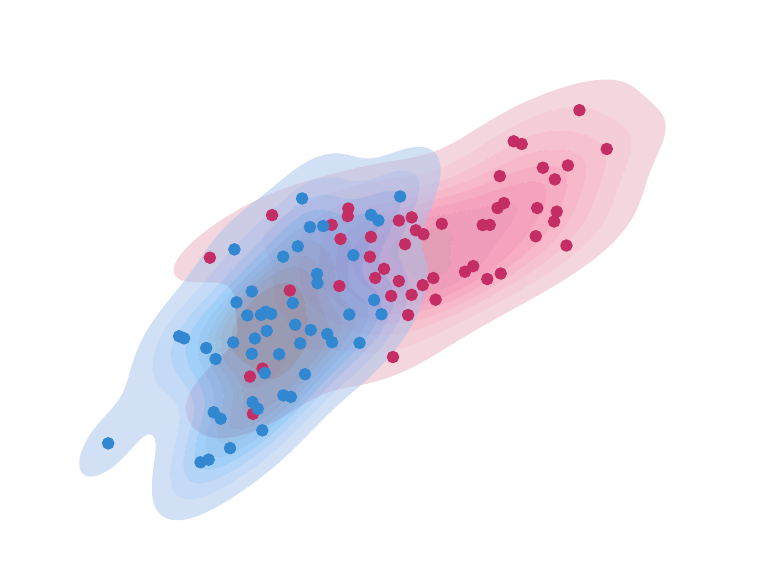}};
      \fill[gray, opacity=0.1] (current bounding box.south west) rectangle (current bounding box.north east);      
  \end{scope}

  \begin{scope}
      \clip[use Hobby shortcut, closed=true]
        (6 + 5.5, -1.75) ..
        (6 + 5.5, 0.75) ..
        (6 + 4.85, 1.5) ..
        (6 + 2.5, 4.1) ..
        (6 + 0, 4) ..
        (6 + -5.15, 3.5) ..
        (6 + -6, 1) ..
        (6 + -4, -3) ..
        (6 + 0, -4) ..
        (6 + 1, -4) ..
        (6 + 3, -3.5) ..
        (6 + 4.25, -2.75);

      \node at ( 1*6, 0) (z2) {\includegraphics[]{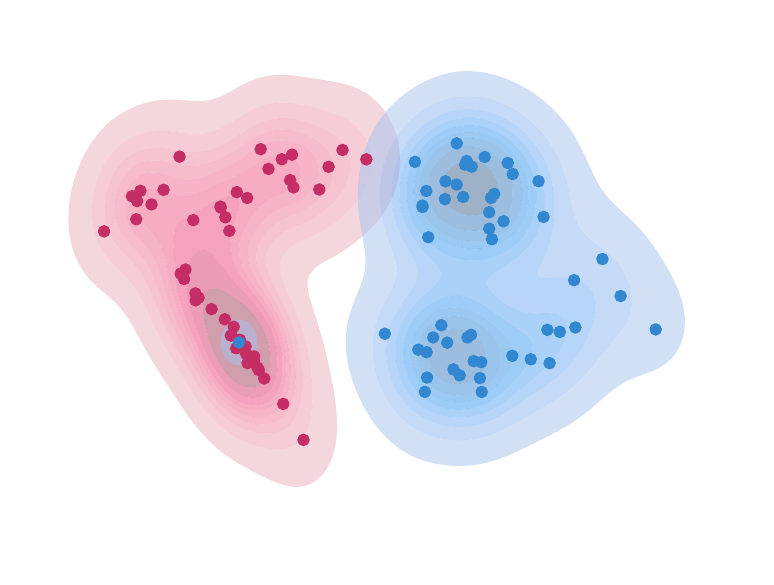}};
      \fill[gray, opacity=0.1] (current bounding box.south west) rectangle (current bounding box.north east);      
  \end{scope}

\draw[->, bend left, line width=0.125cm] (-1.5, 3.55) to node[above, midway, xshift=-12, yshift=6]{\scalebox{3}{$f$}} (2.75, 3.65);
\draw[<-, bend right, line width=0.125cm] (-6.5, -2.15) to node[above, midway, xshift=-12, yshift=6]{\scalebox{3}{$\inv{f}$}} (6.15, -3.25);
  
\node[inner sep=0, outer sep=0] at (-10, 3.5) {\scalebox{5}{\strut $\domain$}};
\node[inner sep=0, outer sep=0] at (11, 3.5) {\scalebox{5}{\strut $\latent$}};

\end{tikzpicture}

%% file: src/02.Background.tex
\section{Background}
\label{sec:background}

We cover related work on concept-based neural networks and normalizing flows.

\subsection{Concept-Based and Prototypical Neural Networks}

Concept-aware models~\cite{poeta2023concept} are designed to produce high-level explanations that are more directly meaningful to human examiners.  
The semantic quality of these explanations is naturally borne by the model's analytic form~\cite{razIMLForm2022,DBLP:conf/xai/PocheHB23} rather than uninformed \textit{post hoc} correlations~\cite{falsifiable2020}, which may not be meaningful~\cite{poursabzi2021manipulating}.
\emph{Fully supervised} approaches to concept learning rely on labeled training data that is fully annotated, indicating for example the presence of bone spurs in medical images or the colors of individual body parts in photographs of birds.
Notable examples include the concept bottleneck model~\cite{koh2020concept} and the concept embedding model~\cite{espinosa2022concept}.
Such approaches are hindered by the cost and constraints of human labor, restricting their use in \emph{unsupervised} settings.
While \emph{weakly supervised} tasks, which don't have annotations, may retain class labels, \emph{fully unsupervised} tasks have neither.
This paper will focus on weakly supervised learning with prototypical neural networks~\cite{li2018deep,protopnet}, the predominant approach in this setting.
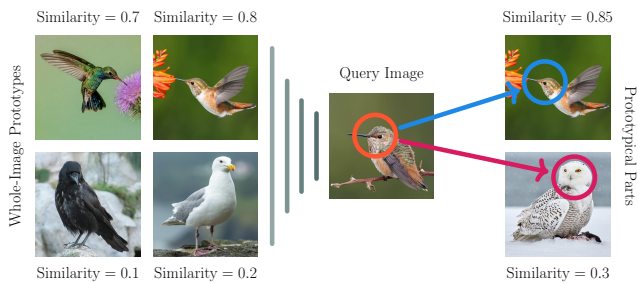
\begin{figure}[t]
  \centering
  \resizebox{!}{3.75cm}{\input{tikz/birby}}
  \caption{Examples of whole-image (left) and prototypical parts (right) explanations.}
  \label{fig:all-proto-expls}
\end{figure}
Though they are related to several other tasks in the literature, including data summarization~\cite{gurumoorthy2019efficient,DBLP:conf/xai/PocheHB23}, example-based post hoc explanation~\cite{filho2023explainable,DBLP:conf/xai/PocheHB23}, and few-shot learning~\cite{snell2017prototypical}, our goal here is to learn prototypes for \emph{intrinsically explainable} case-based reasoning~\cite{li2018deep,protopnet}.

\paragraph{Prototypical Neural Networks}
Prototype networks provide local explanations by relating decisions made on input data to discriminative abstractions called \emph{prototypes}.
In \cref{fig:all-proto-expls}, we illustrate how images can be explained with (a) whole-image prototypes and (b) prototypical parts corresponding to specific portions of an image.
Decisions are made by a composite mapping $\domain \xto{\text{\tiny $f$}} \latent \xto{\text{\tiny $g$}} \classes$ that sorts the data $\data \in \domain$ into mutually exclusive classes $\class \in \classes$.
This first involves learning a transformation $\data \xmapsto{\text{\tiny $f$}} \embedding$ of our Euclidean data space $\domain$ into a latent inner-product space $\latent$ whose parameters specify a set of prototypes related to the class labels.
A classifier $\embedding \xmapsto{\text{\tiny $g$}} \class$ then makes decisions based on a similarity kernel $\kappa: \latent \times \latent \to \mathbb{R}$ imposed on the learned representations.

Critically, the latent representations $\embedding$ don't exist in the same space as our data points $\data$.
So, if $\domain$ is a space of images for example, nothing so far gives us a way to visually understand how images nor prototypes are represented in $\latent$.
This requires an \emph{interpretation} transform $\latent \xto{\text{\tiny $h$}} \domain$ going in the other direction, and several proposals have been offered in the literature.
Li et al.\ proposed an autoencoder framework with $f$ as the encoder and $h$ as the decoder~\cite{li2018deep}.
This, however, is only capable of approximate reconstructions, with visualization fidelity susceptible to shifts in the latent prototypical distributions.
As an alternative, Chen et al.\ proposed \emph{prototype projection}~\cite{protopnet}.
Under this scheme---illustrated in \cref{fig:proto-points}---learned prototype representations are replaced by the image embeddings of their most similar training images.
Although this guarantees an association between certain points in $\latent$ and training data from $\domain$, the fact that $f$ is typically not injective often results in a human-machine semantic similarity gap~\cite{kim2022hive} when multiple disparate data points are associated with the same latent point.
Other approaches choose instead to compute the nearest-neighbor image embeddings to each prototypical point~\cite{ma2024looks,Wan_2024_WACV}.
Point-based approaches such as these have relatively low representation power due to (1) the necessary sparsity of high-dimensional spaces and (2) a dearth of variability measures~\cite{wang2023mixture}.

Instead of prototypical points, we propose learning prototypes as \emph{distributions} over the latent space as shown in \cref{fig:proto-dists}.
Moreover, we recover an intrinsic association between $\latent$ and $\domain$ by learning these distributions \emph{invertibly} with \emph{normalizing flows} (\textit{cf.} \cref{subsec:normFlowsBackground}), providing us an implicit mapping $\inv{f}: \latent \to \domain$ that interprets and visualizes $\latent$ through the eyes of $\domain$.

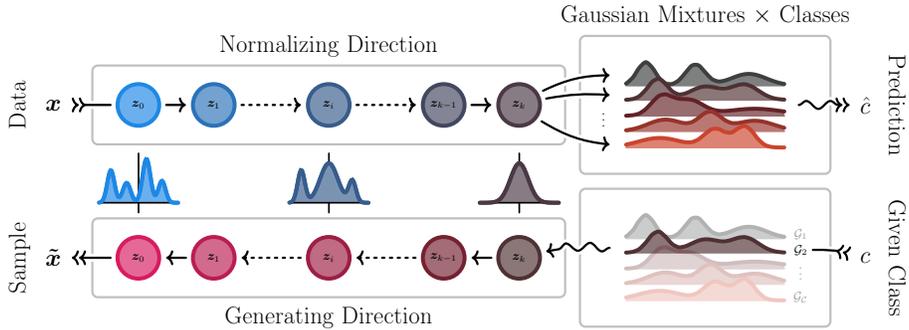
\begin{figure}[t]
  \centering
  \resizebox{\linewidth}{!}{\input{tikz/architecture}}
  \caption{%
    An overview of the proposed \Ours{} model that composes a normalizing flow and Gaussian mixture models.
    In the \emph{normalizing} $\domain \xto{\text{\tiny $f$}} \latent$ direction, the invertible composition $f = f_k \circ \dots \circ f_1$ \emph{pulls back} the structured latent density $\platent$ to the complex data density $\pdata$.
    The \emph{generating} $\domain \xfrom{\text{\tiny $\inv{f}$}} \latent$ direction \emph{generates} points $\tilde{\data} \sim \pdata$ implicitly by \emph{pushing forward} samples  $\tilde{\embedding} \sim \platent$ from the latent distribution along the inverse $\inv{f}$.
  }\label{fig:ours}
\end{figure}

\paragraph{Related Prototypical Neural Networks}
Somewhat similarly, Ma et al.~\cite{ma2024looks} propose learning prototypical balls---latent hyperspheres in $\latent$---rather than prototypical points.
To interpret a given prototype, they visualize every training image whose latent representation lies inside the ball for that prototype.
ProtoSegNet~\cite{gerstenberger2023differentiable} models prototypes as the components of Gaussian mixture models (GMMs) learned over the latent space by a prototypical segmentation network; however, their prototype interpretations are limited to indirect visualizations involving nearest neighbors in the data.
MGProto~\cite{wang2023mixture} again models prototypes as GMMs, this time choosing to visualize prototypes by applying prototype projection to the components' mean points.
Peters~\cite{peters2022extending} attempts to extract prototypical points from the latent space of a generative model using a normalizing flow generator; these points unfortunately look like noise, so nearest neighbors are used instead.
Unlike prior work, we are able to faithfully visualize the full learned prototypical distributions due to our invertible normalizing flow backbone.

\paragraph{Prototypical Parts}
While full-image prototypical neural networks are intrinsically explainable, the part-level explanations of existing prototypical part neural networks are typically considered post hoc~\cite{Carmichael_2024_WACV,PRP,DBLP:conf/iccv/HamdiGG21,sacha2023interpretability,wolf2023keep}.
The exceptions are models like PixPNet~\cite{Carmichael_2024_WACV}, ProtINN~\cite{peters2022extending}, and ProtoBBNet~\cite{gerstenberger2023differentiable}, which either impose constraints on the sizes of their receptive fields or process images by patches (using super-pixels or pre-defined grids, for example).
See these papers~\cite{PRP,wolf2023keep,protopnet} for more details on post hoc part-level explainability in prototype networks.

\subsection{Normalizing Flows}\label{subsec:normFlowsBackground}
A normalizing flow is an unsupervised density estimator $\model: \domain \to \latent$ that \emph{invertibly} transforms between the data and latent spaces.
Empirical information from data \emph{flows} to the latent space through $f$, and density inferences are recovered by the inverse $\inv{f}: \latent \to \domain$.
Given a data distribution $\pdata$ and a latent distribution $\platent$ over $\domain$ and $\latent$ respectively, the impact of changing variables~\cite{papamakarios2021normalizing} on a random variable $\random{x} \sim \pdata$ via $f(\random{x}) = \random{z}$ is given by the following formula.
\begin{equation}\label{eq:normFlow}
    \pdata(\random{x}) = \platent(f(\random{x})) \cdot \left| \det\left(\frac{\partial f} {\partial \random{x}}\right) \right|
\end{equation}
This allows us to formally specify the unknown data distribution $\pdata$ as the \emph{pushforward} of a latent distribution $\platent$, with the inverse transformation $\inv{f}$ responsible for \emph{``pushing''} the latent density $\platent$, which is typically taken to be well-behaved, \emph{``forward''} onto the more unwieldy data distribution $\pdata$.
With full knowledge and control over $\platent$, we can leverage $\inv{f}$ as a generative model for sampling from $\pdata$ implicitly through the transformation $\inv{f}(\random{z}) \sim \pdata$ where $\random{z} \sim \platent$.
We call this model a ``normalizing flow'' because $f$ \emph{``normalizes''} the complicated data distribution by \emph{``flowing''} information into the latent space~\cite{papamakarios2021normalizing}.

The mapping $f$ is typically implemented as a composition of invertible functions learned by neural networks.
A neural network can be made invertible by constructing a discrete- or continuous-time flow, such as coupling flows, autoregressive flows, linear flows, and planar flows~\cite{kobyzev2020normalizing}.
The model can then be trained by maximum likelihood estimation as in \cref{eq:normFlow}.

\subsection{Joint Generative and Predictive Modeling}

\paragraph{Normalizing Flows}

Invertible neural networks have already seen use for classification tasks~\cite{behrmann2019invertible,mangalam2022reversible,etmann2020iunets,DBLP:conf/iclr/JacobsenSO18} partly due to their memory efficiency during training.
However, our focus here is on joint generative and predictive modeling.
Fetaya et al.~\cite{DBLP:conf/iclr/FetayaJGZ20} composed a normalizing flow (Glow~\cite{glow}) with a latent GMM to perform joint generation and prediction.
Atanov et al.~\cite{atanov2019semi} and Izmailov et al.~\cite{izmailov2020semi} take similar approaches but with alternative normalizing flows~\cite{DBLP:conf/iclr/DinhSB17,DBLP:conf/iclr/GrathwohlCBSD19} and extensions to semi-supervised learning.
Ardizzone et al.~\cite{NEURIPS2020_593906af,mackowiak2021generative} continue this trend by adding an information bottleneck term to their loss that balances between performance and robustness.
Rather than training a separate classifier (as with \emph{hybrid} models), these approaches classify by estimating the latent-conditional class likelihood $\platent(y \mid \embedding)$.
Although our approach happens to overlap with these models, we are motivated directly by intrinsically explainable modeling using \emph{prototypes.}
Notably, we choose to learn \emph{multiple} prototypical distributions in the latent space as the base for a normalizing flow, allowing for potentially greater mode coverage~\cite{zhong2019rethinking}.

\paragraph{Other Generative Approaches}

Both hybrid and joint generative/predictive models have been based on other generating methods than normalizing flows.
Some notable hybrid models have used deep belief networks~\cite{hinton2007recognize,ng2001discriminative}, autoencoders~\cite{he2022masked}, generative adversarial models (GANs)~\cite{jabbar2021survey}, and diffusion models~\cite{cao2024survey,yang2024diffusionsurvey}.
Closely related, some supervised tasks involve generating output that doesn't match the input data distribution.
Notable examples here---based on GANs~\cite{jabbar2021survey} and diffusion models~\cite{cao2024survey,yang2024diffusionsurvey}---have been applied to object detection~\cite{ehsani2018segan}, image classification~\cite{yang2023diffmic}, forecasting~\cite{yan2021scoregrad}, and next-frame video prediction~\cite{yang2023diffusion}.
Joint generative/predictive modeling dates back to 1996 when Revow et al.\ proposed generative models built from B-splines for recognizing and generating handwritten digits~\cite{revow1996gendigits}.
More recent approaches have involved variational autoencoders~\cite{DBLP:conf/iclr/SchottRBB19}, diffusion models~\cite{DBLP:conf/nips/HanZZ22,DBLP:conf/iccv/LiPDBP23}, and scored-based generative classifiers~\cite{zimmermann2021score}.
Unfortunately, none of these generators provide an exact likelihood, an important consideration for explainability.
They also fail to enforce a faithful association between the latent and data distributions---only providing approximate inverse mappings from latent to data space---unlike the exact inverse $\inv{f}: \latent \to \domain$ granted by normalizing flows (see \cref{fig:teaser}).
We therefore only consider normalizing flow generators.

%% file: tikz/birby.tex
\begin{tikzpicture}[ultra thick, line cap=round]
  \node at (0*2.25,  0*2.25) (b1) {\includegraphics[width=4cm]{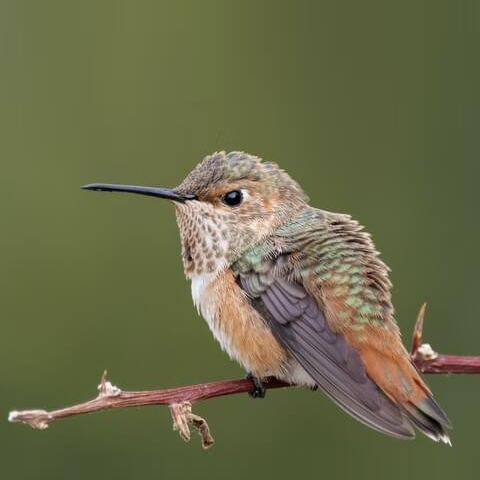}};
  \node at (-3*2.25,  1*2.25) (b2) {\includegraphics[width=4cm]{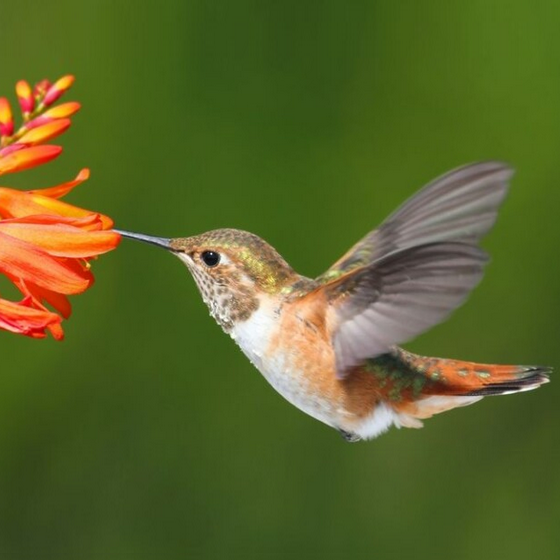}};
  \node at (-3*2.25, -1*2.25) (b3) {\includegraphics[width=4cm]{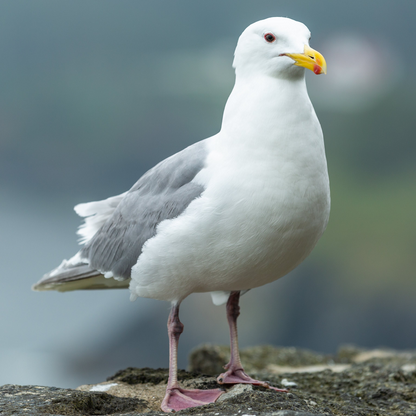}};
  \node at (-5*2.25,  1*2.25) (b4) {\includegraphics[width=4cm]{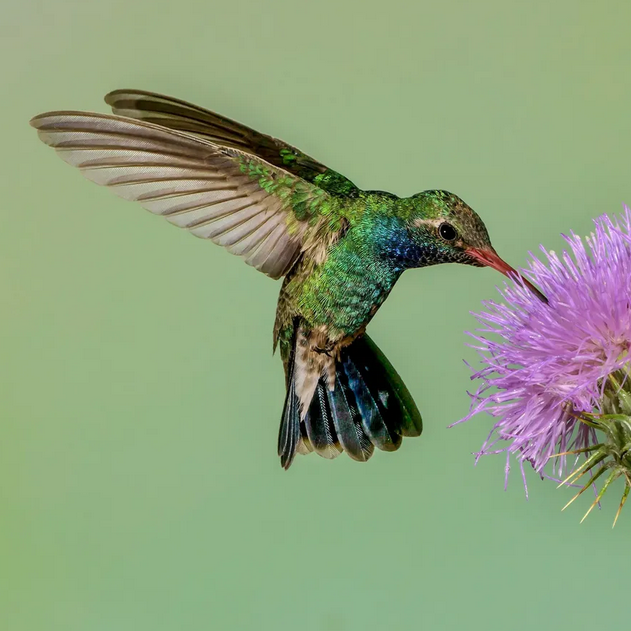}};
  \node at (-5*2.25, -1*2.25) (b5) {\includegraphics[width=4cm]{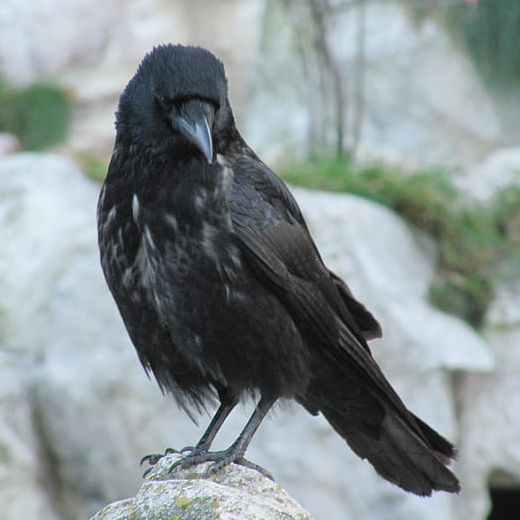}};

  \draw[-, line width=0.17
5cm, stone, opacity=0.8] (-2.5, -1.25) to (-2.5, 1.25);
  \draw[-, line width=0.175cm, stone, opacity=0.7] (-3.075, -1.75) to (-3.075, 1.75);
  \draw[-, line width=0.175cm, stone, opacity=0.6] (-3.625, -2.5) to (-3.625, 2.5);
  \draw[-, line width=0.175cm, stone, opacity=0.5] (-4.2, -3.75) to (-4.2, 3.75);

  \node at (0, 2.75) {\LARGE \strut Query Image};
  \node[above=0 of b2] {\LARGE \strut $\text{Similarity} = 0.8$};
  \node[below=0 of b3] {\LARGE \strut $\text{Similarity} = 0.2$};
  \node[above=0 of b4] {\LARGE \strut $\text{Similarity} = 0.7$};
  \node[below=0 of b5] {\LARGE \strut $\text{Similarity} = 0.1$};
  \node[rotate=90] at (-6*2.25 - 0.5, 0) {\LARGE \strut Whole-Image Prototypes};

  \node at (3*2.25,  1*2.25) (b2) {\includegraphics[width=4cm]{figs/birby/birby1.png}};
  \node at (3*2.25, -1*2.25) (b3) {\includegraphics[width=4cm]{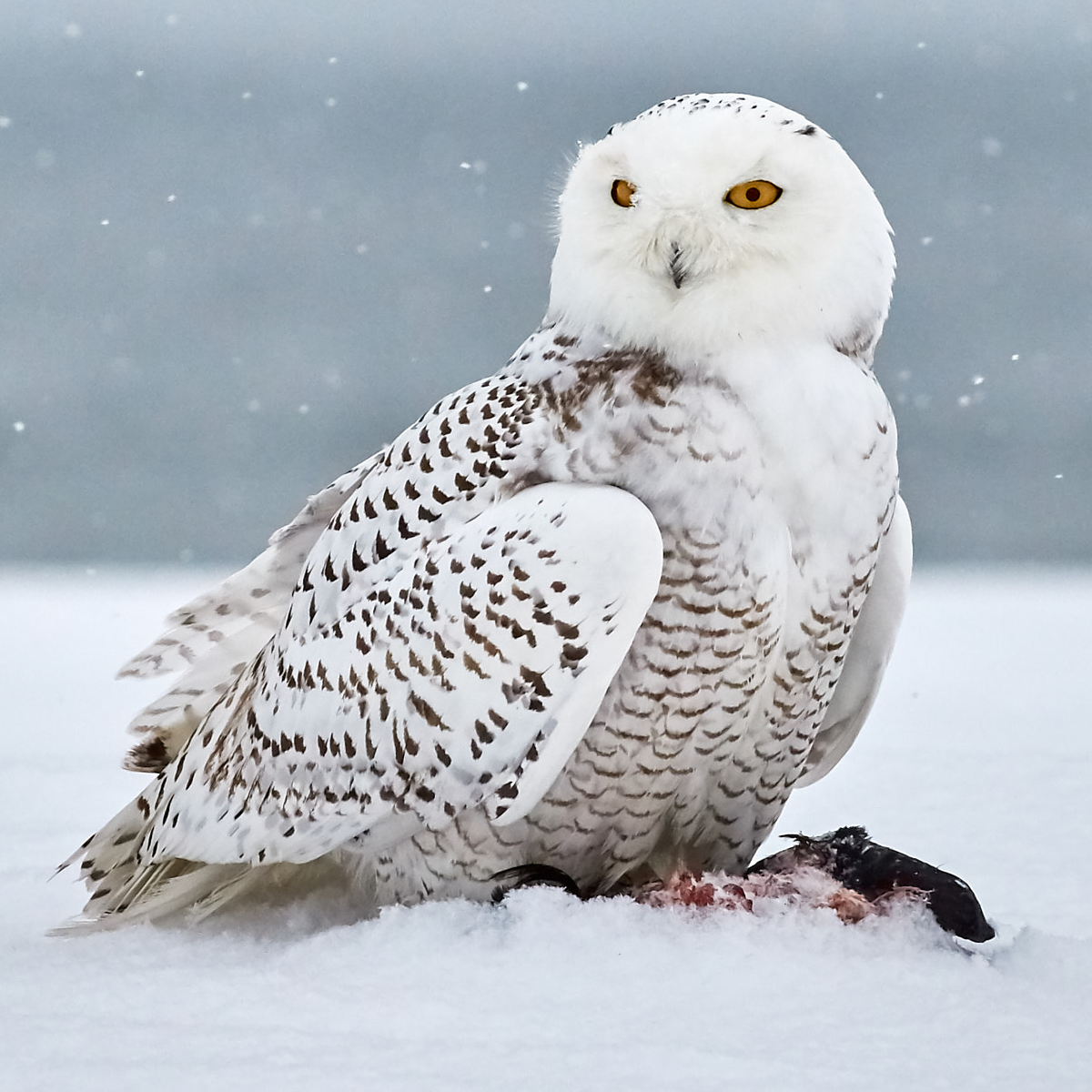}};

  \node[circle, opacity=0, draw=foreground, line width=0.175cm, inner sep=16, outer sep=6] at ($(b1) + (-0.25, 0.4)$) (c0) {};

  \node[circle, draw=orange, line width=0.175cm, inner sep=16, outer sep=6] at ($(b1) + (-0.25, 0.4)$) (c1) {};

  \node[circle, draw=blue, line width=0.175cm, inner sep=16, outer sep=6] at ($(b2) + (-0.5, 0.2)$) (c2) {};

  \node[circle, draw=red, line width=0.175cm, inner sep=16, outer sep=6] at ($(b3) + (0.625, 1.05)$) (c3) {};

  \draw[->, blue, line width=0.175cm] (c0) to (c2);
  \draw[->, red, line width=0.175cm] (c1) to (c3);

  \node[above=0 of b2] {\LARGE \strut $\text{Similarity} = 0.85$};
  \node[below=0 of b3] {\LARGE \strut $\text{Similarity} = 0.3$};
  \node[rotate=270] at (4*2.25 + 0.5, 0) {\LARGE \strut Prototypical Parts};
\end{tikzpicture}

%% file: tikz/architecture.tex
\begin{tikzpicture}[ultra thick, line cap=round, node distance=2,
                    znode/.style={circle, fill opacity=1, text opacity=1, minimum size=30, inner sep=0, outer sep=5, line width=0.1cm}]
  \scoped[]{
    \node[znode, draw=blue, fill=blue!66!background] at (0, 0) (z0) {$\embedding_0$};
    \node[znode, draw=stone!33!pumice!25!blue, fill=stone!33!pumice!25!blue!66!background, right=1*0.5 of z0] (z1) {$\embedding_1$};
    \node[znode, draw=stone!33!pumice!50!blue, fill=stone!33!pumice!50!blue!66!background, right=3*0.5 of z1] (zi) {$\embedding_i$};
    \node[znode, draw=stone!33!pumice!75!blue, fill=stone!33!pumice!75!blue!66!background, right=3*0.5 of zi] (zkm1) {$\embedding_{k-1}$};
    \node[znode, draw=stone!33!pumice!90!blue, fill=stone!33!pumice!90!blue!66!background, right=1*0.5 of zkm1] (zk) {$\embedding_k$};

    \node[outer sep=4.5] at ($(z0) + (-2.15, 0)$) (x) {\LARGE $\data$};
    \node[rotate=90] at ($(x) + (-0.85, 0)$) (xhat) {\thead{\LARGE Data}};

    \draw[-, line width=0.2cm, draw=background] (x) to (z0);
    \draw[>.>-] (x) to (z0);

    \draw[->] (z0) to node[above=0.333cm, midway] {} (z1);
    \draw[->, loosely dotted] (z1) to node[above=0.333cm, xshift=-0.75cm] {} node[above=0.5cm] {} node[above=0.333cm, xshift=0.75cm] {} (zi);
    \draw[->, loosely dotted] (zi) to node[above=0.333cm, xshift=-0.75cm] {} node[above=0.5cm] {} node[above=0.333cm, xshift=0.75cm] {} (zkm1);
    \draw[->] (zkm1) to node[above=0.333cm, midway] {} (zk);

    \coordinate (gg) at ($(zk.east) + (4, 0)$);

    \node at ($(gg) + (0,0.8)$) (g1) {\gaussian{2}{-1.5}{0.225}{-0.25}{0.25}{1.1}{0.45}{pumice!33!stone}};
    \node at ($(gg) + (0,0.4)$) (g3) {\gaussian{2}{-1.2}{0.25}{0}{0.4}{1}{0.35}{pumice!66!stone}};
    \node at ($(gg) + (0, 0)$) (g6) {\gaussian{2}{-1.1}{0.275}{-0.4}{0.35}{1.25}{0.75}{pumice}};
    \node at ($(gg) + (0,-0.4)$) (g8) {\gaussian{2}{-1}{0.45}{0.12}{0.285}{1.25}{0.5}{pumice!66!orange}};
    \node at ($(gg) + (0,-0.8)$) (g9) {\gaussian{2}{-1.15}{0.55}{0.25}{0.285}{1}{0.25}{pumice!33!orange}};

    \draw[-, line width=0.2cm, draw=background, bend left=12.5 ] (zk) to ($(g1.west) + (-0.25, -0.05)$);
    \draw[-, line width=0.2cm, draw=background, bend left=7    ] (zk) to ($(g3.west) + (-0.25, -0.15)$);
    \draw[-, line width=0.2cm, draw=background, bend left=-12.5] (zk) to ($(g9.west) + (-0.25, -0.2)$);

    \draw[->, bend left=12.5  ] (zk) to ($(g1.west) + (-0.25, -0.05)$);
    \draw[->, bend left=7     ] (zk) to ($(g3.west) + (-0.25, -0.15)$);
    \draw[->, bend left=-12.5 ] (zk) to ($(g9.west) + (-0.25, -0.2)$);

    \node at ($(g6.west) + (-0.4, -0.27)$) (vdots) {$\vdots$};

    \node[outer sep=4.5] at ($(zk) + (8.75, 0)$) (y) {\LARGE $\pred$};
    \node[rotate=270] at ($(y) + (0.75, 0)$) (yhat) {\thead{\LARGE Prediction}};

    \draw[-, line width=0.2cm, wavy, draw=background] ($(g6.east) + (0.2, 0)$) to (y);
    \draw[->.>, wavy] ($(g6.east) + (0.2, 0)$) to (y);

    \scoped[on background layer]{
      \draw[ultra thick, rounded corners, line cap=round, draw opacity=0.25, fill opacity=0]
        ($(z0.north west) + (-0.666, 0.5)$) rectangle ($(zk.south east) + (0.666, -0.5)$);
      \node at ($(z0.north west)!0.5!(zk.north east) + (0, 1)$) {\thead{\LARGE Normalizing Direction}};
    }

    \scoped[on background layer]{
      \draw[ultra thick, rounded corners, line cap=round, draw opacity=0.25, fill opacity=0]
        ($(g1.north west) + (-1, 0.5)$) rectangle ($(g9.south east) + (1, -0.5)$);
      \node at ($(g1.north west)!0.5!(g1.north east) + (0, 1)$) {\thead{\LARGE Gaussian Mixtures $\times$ Classes}};
    }

    \node[outer sep=0, inner sep=0, below=0.5 of z0, xshift=-0.05cm] (p0) {\distro[90]{0}{1}{1}{blue}};
    \node[outer sep=0, inner sep=0, below=0.5 of zi, xshift=-0.05cm] (pi) {\distro[70]{1}{1}{0}{stone!33!pumice!50!blue}};
    \node[outer sep=0, inner sep=0, below=0.5 of zk, xshift=-0.05cm] (pk) {\distro{1}{0}{0}{stone!33!pumice!90!blue}};
  }

  \scoped[yshift=-110]{
    \node[znode, draw=red, fill=red!66!background] at (0, 0) (z0) {$\embedding_0$};
    \node[znode, draw=red!75!pumice, fill=red!75!pumice!66!background, right=1*0.5 of z0] (z1) {$\embedding_1$};
    \node[znode, draw=red!50!pumice, fill=red!50!pumice!66!background, right=3*0.5 of z1] (zi) {$\embedding_i$};
    \node[znode, draw=red!15!pumice, fill=red!15!pumice!66!background, right=3*0.5 of zi] (zkm1) {$\embedding_{k-1}$};
    \node[znode, draw=pumice!66!stone, fill=pumice!66!stone!66!background, right=1*0.5 of zkm1] (zk) {$\embedding_k$};

    \node[outer sep=4.5] at ($(z0) + (-2.15, 0)$) (x) {\LARGE $\gen$};
    \node[rotate=90] at ($(x) + (-0.85, 0)$) (xhat) {\thead{\LARGE Sample}};

    \draw[-, line width=0.2cm, draw=background] (z0) to (x);
    \draw[->.>] (z0) to (x);

    \draw[<-] (z0) to node[above=0.333cm, midway] {} (z1);
    \draw[<-, loosely dotted] (z1) to node[above=0.333cm, xshift=-0.75cm] {} node[above=0.5cm] {} node[above=0.333cm, xshift=0.75cm] {} (zi);
    \draw[<-, loosely dotted] (zi) to node[above=0.333cm, xshift=-0.75cm] {} node[above=0.5cm] {} node[above=0.333cm, xshift=0.75cm] {} (zkm1);
    \draw[<-] (zkm1) to node[above=0.333cm, midway] {} (zk);

    \coordinate (gg) at ($(zk.east) + (4, 0)$);

    \scoped[transparency group, opacity=0.4]{
      \node at ($(gg) + (0,0.8)$) (g1) {\gaussian{2}{-1.5}{0.225}{-0.25}{0.25}{1.1}{0.45}{pumice!33!stone}};
    }
    \node at ($(gg) + (0,0.4)$) (g3) {\gaussian{2}{-1.2}{0.25}{0}{0.4}{1}{0.35}{pumice!66!stone}};
    \scoped[transparency group, opacity=0.3]{
      \node at ($(gg) + (0, 0)$) (g6) {\gaussian{2}{-1.1}{0.275}{-0.4}{0.35}{1.25}{0.75}{pumice}};
      \node at ($(gg) + (0,-0.4)$) (g8) {\gaussian{2}{-1}{0.45}{0.12}{0.285}{1.25}{0.5}{pumice!66!orange}};
      \node at ($(gg) + (0,-0.8)$) (g9) {\gaussian{2}{-1.15}{0.55}{0.25}{0.285}{1}{0.25}{pumice!33!orange}};
    }

    \draw[line width=0.2cm, draw=background, wavy, bend right=7] ($(g3.west) + (-0.25, -0.15)$) to (zk);
    \draw[->, wavy, bend right=7] ($(g3.west) + (-0.25, -0.15)$) to (zk);

    \scoped[on background layer]{
      \draw[ultra thick, rounded corners, line cap=round, draw opacity=0.25, fill opacity=0]
        ($(z0.north west) + (-0.666, 0.5)$) rectangle ($(zk.south east) + (0.666, -0.5)$);
      \node at ($(z0.south west)!0.5!(zk.south east) + (0, -1)$) {\thead{\LARGE Generating Direction}};
    }

    \scoped[on background layer]{
      \draw[ultra thick, rounded corners, line cap=round, draw opacity=0.25, fill opacity=0]
        ($(g1.north west) + (-1, 0.5)$) rectangle ($(g9.south east) + (1, -0.5)$);
    }
    \node[outer sep=4.5] at ($(zk) + (8.75, 0)$) (c) {\LARGE $\class$};
    \node[rotate=270] at ($(c) + (0.75, 0)$) (chat) {\thead{\LARGE Given Class}};

    \node[opacity=0.4] at ($(g1.east) + (0.25, -0.2)$) (G1) {$\mathcal{G}_{1}$};
    \node at ($(g3.east) + (0.25, -0.2)$) (G2) {$\mathcal{G}_{2}$};
    \node[opacity=0.4] at ($(g8.east) + (0.25, 0.1)$) (G4) {$\vdots$};
    \node[opacity=0.4] at ($(g9.east) + (0.25, -0.2)$) (G5) {$\mathcal{G}_{\mathcal{C}}$};

    \draw[-, line width=0.2cm, draw=background, bend right=7] (c) to (G2);
    \draw[>.>-, bend right=7] (c) to (G2);
  }
\end{tikzpicture}

%% file: src/03.Methodology.tex
\section{Invertible Prototypical Networks}
\label{sec:methods}

We propose an intrinsically interpretable approach to learning that bridges together generative classification with concept-based  networks---two key ideas from the explainable AI and generative AI literature.
As described in \cref{sec:background}, we aim to invertibly learn latent prototypical distributions.
In this section, we first describe the normalizing-flow backbone of our architecture.
Then, we detail how prototypical distributions are learned over the latent space.
Finally, we discuss our training methodology for \Ours{}, our proposed architecture for joint predictive and generative modeling.
An overview of \Ours{} is given in \cref{fig:ours}.

\paragraph{Normalizing Flow Backbone}

We base our normalizing flow on DenseFlow~\cite{grcic2021densely}, a state-of-the-art unconditional density estimator~\cite{denseflowsota}.
DenseFlow is comprised of invertible Glow-like~\cite{glow} modules using cross-unit coupling and densely connected blocks fused with Nystr\"{o}m self-attention,
increasing expressiveness by incrementally augmenting latent vectors with noise.
We replace DenseFlow's \emph{unconditional} latent distribution with \emph{conditional} distributions as follows.

\paragraph{Prototypical Gaussian Mixture Classifier}\label{subsec:gmm}
Whereas other prototypical neural networks usually learn latent \emph{points} to represent their prototypes, our prototypes are given by \emph{probability distributions} learned over the latent space.
By leveraging the inverse mapping $\inv{f}: \latent \to \domain$, these can be reinterpreted as probability distributions over data, providing a direct, faithful, and accurate visualizations of learned prototypes.
For each class $c \,{\in}\, \classes \,{=}\, \set{1, \dots, C}$, we specify a $K$-component GMM $\proto_c$ whose components represent prototypical distributions, so that each class has $K$ associated prototypical distributions.
The components are weighted by $\weight_c \,{=}\, (\pi_{c, 1}, \dots, \pi_{c, K})$.
\begin{equation}\label{eq:prototypes}
  \proto_c
  = \sum_{k = 1}^{K} \pi_{c, k} \proto_{c, k}
  = \sum_{k = 1}^{K} \pi_{c, k} \normal(\mean_{c, k}, \covariance_{c, k})
\end{equation}
The class-conditional likelihood is given below, where $\sigma$ is the softmax function.
\begin{equation}\label{eq:gmmClass}
  \platent(\embedding \mid y) = \sum_{k=1}^{K} \softmax(\pi_{y, k})\, \normal(\embedding \,;\, \mean_{y, k}, \covariance_{y, k})
\end{equation}
As is often done~\cite{gepperth2021gradient}, we constrain $\proto_c$ by (1) asserting $\covariance_{c, k}$ are diagonal, (2) clipping $\covariance_{c, k}$ above zero, and (3) enforcing $\weight_c^\trans\weight_c = 1$.
Applying Bayes' theorem to \cref{eq:gmmClass}, we derive the expression below for the data-conditional class likelihood.
\begin{equation}\label{eq:gmmTotal}
  \pdata(y \mid \data) = \frac{%
    \sum_{k=1}^{K} \softmax(\pi_{y,k})\, \normal(f(\data) \:;\: \mean_{y,k}, \covariance_{y,k})
  }{%
    \sum_{c=1}^{C} \sum_{k=1}^{K} \softmax(\pi_{c,k})\, \normal(f(\data) \,;\, \mean_{c,k}, \covariance_{c,k})
  }
\end{equation}
Relating this back to our discussion in \cref{sec:background}, we can induce the similarity kernel $\kappa$ at a given point $\embedding \in \latent$ by looking at the mean $\mean_{c, k}$ of the prototype that maximizes the class-conditional likelihood $\kappa(\embedding, \mean_{c, k}) = \platent(\embedding \mid y=c; \pi_{c,k}, \mean_{c,k}, \covariance_{c,k})$.
The classifier $g$ can be written $g(f(\embedding)) = \argmax_{y \in \classes}\pdata(y \mid f(\embedding))$ using \cref{eq:gmmTotal}. 
Finally, and most interestingly, our interpretation function $h$ is simply the inverse transform $h = \inv{f}$ induced by the normalizing flow.
This imposes no additional constraints on the learned prototypes, unlike prior work~\cite{protopnet,ma2024looks,Wan_2024_WACV,li2018deep}.

\paragraph{Training}\label{subsec:training}
We train our model to maximize the categorical cross entropy, denoted $\loss_{\ce}$, with auxiliary loss terms.
Since it is known to improve model robustness, we adapt the proposed \emph{consistency regularization} loss from~\cite{izmailov2020semi}.
This encourages the model to be \emph{invariant} to certain perturbations or augmentations of the training data by penalizing the model for predicting two different classes $\dot{y} \neq \ddot{y}$ for two perturbations $\dot{\data}$ and $\ddot{\data}$ of the same data point $\data \in \domain$, expressed below.
\begin{equation}\label{eq:Lconsistency}
  \loss_\con(\dot{\data}, \ddot{\data})
  = -\log \pdata(\dot{\data} \mid \ddot{y})
  = -\log \platent\big( f(\dot{\data}) \mid y
  = \ddot{y} \big) - \log \left| \det \left( \frac{\partial f}{\partial \dot{\data}}\right) \right|
\end{equation}
To help support diversity and reduce information overlap between prototypes in each class, we penalize the components within each mixture based on their squared Hellinger distance $\hell$.
However, high-dimensional spaces' asymptotic sparsity---sometimes referred to as the \emph{curse of dimensionality}---limits the usefulness of $\hell$: as the expected discernibility between points decreases, gradients tend to vanish.
To mitigate this, we propose a \emph{modified} divergence $\hellmod$ rescaled based on the embedding dimension $d = \text{dim}(\latent)$.
Given two multivariate Gaussians $\normal_1 = \normal(\mean_1, \covariance_1)$ and $\normal_2 = \normal(\mean_2, \covariance_2)$, the modified divergence $\hellmod(\normal_1, \normal_2)$ between them---which is not a metric---is given below~\cite{llorente2006statistical}.
\begin{equation}\label{eq:hell}
  1 - \frac{
    \det(\covariance_1\covariance_2)^{\sfrac{1}{4d}}
  }{%
    \det\left(
      \frac{\covariance_1 + \covariance_2}{2}
    \right)^{\sfrac{1}{2d}}
  }
  \exp\left(
    -\frac{1}{8d}
    (\mean_1 - \mean_2)^\mathsf{T}
    \left(
      \frac{\covariance_1 + \covariance_2}{2}
    \right)^{-1} \hspace{-1ex}(\mean_1 - \mean_2)
  \right)
\end{equation}
This is bounded within $[0, 1] \subseteq \mathbb{R}$.
The symmetry of \cref{eq:hell} can be leveraged to avoid repetitive computations, yielding the following form for the diversity loss.
\begin{equation}\label{eq:Ldiversity}
  \loss_\div \left( \proto \right)
  = \frac{-2}{C K (K-1)}\sum_{c=1}^{C}\sum_{i=1}^{K-1} \sum_{j=i}^{K} \hellmod(\proto_{c, i}, \proto_{c, j})
\end{equation}
With hyperparameters $\lambda_\con$ and $\lambda_\div$, the final training objective is given below.
\begin{equation}\label{eq:Ltotal}
  \loss = \loss_\ce\big( \pdata(y \mid \data), y \big)
        + \lambda_\con \loss_\con(\dot{\data}, \ddot{\data})
        + \lambda_\div \loss_\div(\proto)
\end{equation}
The collection of mixtures $\proto$ can be trained either by stochastic gradient descent~(SGD) or expectation maximization~(EM).
Empirically, however, we find that SGD results in better classification performance than EM.
We explore $K$-means as an initialization strategy of the means of Gaussians.
We initialize the $K$ mean parameters of the mixture $\proto_c$ responsible for modeling $\platent(\embedding \mid y=c)$ with the $K$-means clusters of the embeddings $\set{f(\data) \suchthat (\data, c) \in (\domain, \classes)}$.
During training, we keep an exponential moving average~(EMA) of the model parameters to reduce training time.
This is a variant of Polyak averaging~\cite{polyak1992acceleration}.

\begin{figure}[t]
  \captionsetup[subfigure]{aboveskip=-1.75ex}
  \centering
  \begin{subfigure}{.535\linewidth}
    \centering
    \input{tikz/protolikely}
    \caption{\strut Queries, prototype means, and most likely parts.}
    \label{subfig:query}
  \end{subfigure}%
  \begin{subfigure}{.2\linewidth}
    \centering
    \input{tikz/protobirb.tex}
    \caption{\strut Mean point.}\label{subfig:mean-cropped}
  \end{subfigure}%
  \begin{subfigure}{.249\linewidth}
    \centering
    \input{tikz/protobirbs.tex}
    \caption{\strut Similar data points.}\label{subfig:cifar-examples}
  \end{subfigure}
  \caption{%
    \hyperref[subfig:query]{(a)}\emph{``This looks like that''}-style explanations of \texttt{bird} (top row) and \texttt{automobile} (bottom row) image classification decisions.
    Rather than using training samples as prototypes, \Ours{} learns prototype distributions directly over the latent space, leading to the \emph{``bird/car-adjacent''} images in the third column.
    The fourth column shows the most-likely dataset image part for each prototypical distribution.
    \hyperref[subfig:mean-cropped]{(b)} The mean point image of the \texttt{bird} prototype with a bird-like figure segmented from the background.
    \hyperref[subfig:cifar-examples]{(c)} Human-picked images from CIFAR-10 that qualitatively match this prototype image.
  }
  \label{fig:cifar10-proto-parts}
\end{figure}
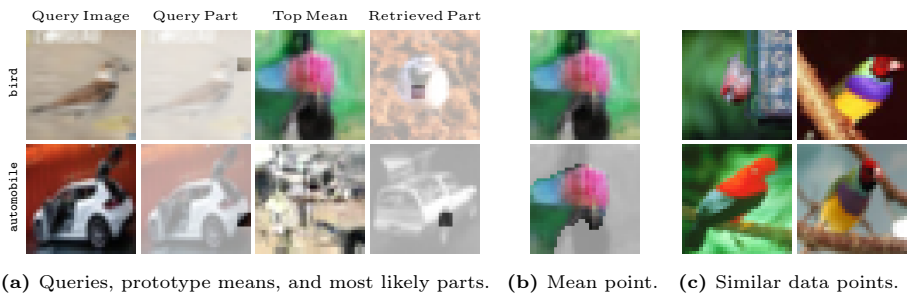

\paragraph{Pruning}
Not all prototypes may end up being useful: they may be redundant, noisy, semantically meaningless, or contribute negligibly to prediction performance.
Pruning such prototypes not only reduces not only the number of parameters in the model but also the quantity of information a user must digest to parse an explanation.
We propose pruning prototypes based on the weights $\pi_{c,k}$ learned on their respective mixture components.
Prototypes with weights below a minimal threshold $\varepsilon$---selected using Otsu's method~\cite{otsuMethod}---are discarded.

\paragraph{Prototypical Parts}
We propose a new approach to explaining prototypical parts.

The original prototypical part network ProtoPNet~\cite{protopnet} enforces a dimensional correspondence between $\domain$ and $\latent$.
They find prototypical parts by first computing an element-wise similarity between each embedded image $\embedding$ of shape $d \times \ell' \times w'$ and a given prototypical point $\mat{p}_k$ that lives in a $d$-dimensional affine subspace of $\latent$, resulting in a similarity map for each prototype.
These maps get upsampled to the image dimension $\ell \times w$ using bicubic interpolation.
The top 5\% scoring pixels in the original image are then finally selected as prototypical parts.

We take a different approach.
Let $\random{x} \sim \pdata$ be a random variable distributed according to our data-generating density, and let $\data \in \set{\data_1, \dots \data_N} \subseteq \domain$ be a query image.
We introduce the notation $\ppart{i}{i+\ell}{j}{j+w}{\data}$ to mean the block of pixels from the image $\data$ with coordinates ranging $\set{i, \dots i+\ell}$ horizontally and $\set{j, \dots j+w}$ vertically.
We then compute a heatmap $\mathcal{M}_{c, k}$ for each prototype $\proto_{c, k}$ as follows.
\vspace{-2.5ex}  %
\begin{align}
  \bar{\data}                                                & \;\coloneq\;   \expval{\random{x}} = \textstyle\frac{1}{N}\sum_{i = 1}^{N} \data_i \label{eq:line1}\\
  \ppart{i}{i+\ell}{j}{j+w}{\bar{\data}}                     & \;\leftarrow\; \ppart{i}{i+\ell}{j}{j+w}{\data} \label{eq:line2}\\
  \ppart{i}{i+\ell}{j}{j+w}{\left(\mathcal{M}_{c, k}\right)} & \;\leftarrow\; \platent\left( f\left( \bar{\data} \right) \mid \proto_{c, k} \right) \label{eq:line3}
\vspace{2.5ex}  %
\end{align}
On line~\eqref{eq:line1}, we begin with an image $\bar{\data}$ computed as the pixel-wise average of every image $\data_i$ in the training data.
On line~\eqref{eq:line2}, we replace the image patch in $\bar{\data}$ specified by the rectangle with lower-left corner at $(i, j)$ and upper-right corner at $(i{+}\ell, j{+}w)$ with the corresponding patch of pixels from a given query image $\data$, leaving the rest of $\bar{\data}$ as it was.
This results in an image whose background is averaged across the entire dataset containing a single patch from a query image.
On line~\eqref{eq:line3}, we then evaluate $\platent( f( \bar{\data} ) \mid \proto_{c, k} )$ and replace the corresponding patch in $\mathcal{M}_{c, k}$ with the prototype-conditional likelihood of this patched image.
Like ProtoPNet, our network's receptive field is the full input, so it would be inappropriate to attribute a subset of pixels to a particular prototype.
However, we can emphasize the top 5\% of pixels to indicate the most influential image parts on a classification decision.
Since we assume independence between image patches, this is an approximate, post hoc method in the same vein as ProtoPNet.

%% file: tikz/protolikely.tex
\begin{tikzpicture}
  \node at (0*1.525,  0*1.525) {\includegraphics[width=1.45cm]{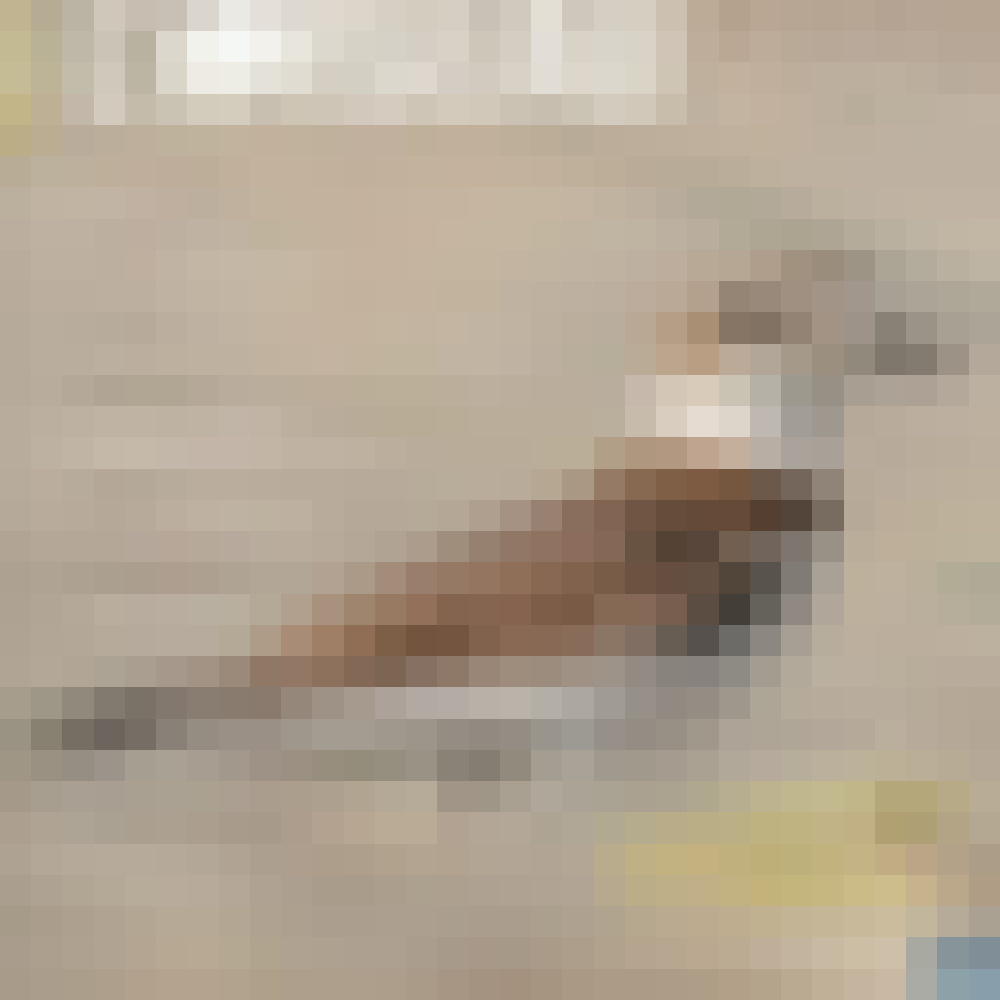}};
  \node at (1*1.525,  0*1.525) {\includegraphics[width=1.45cm]{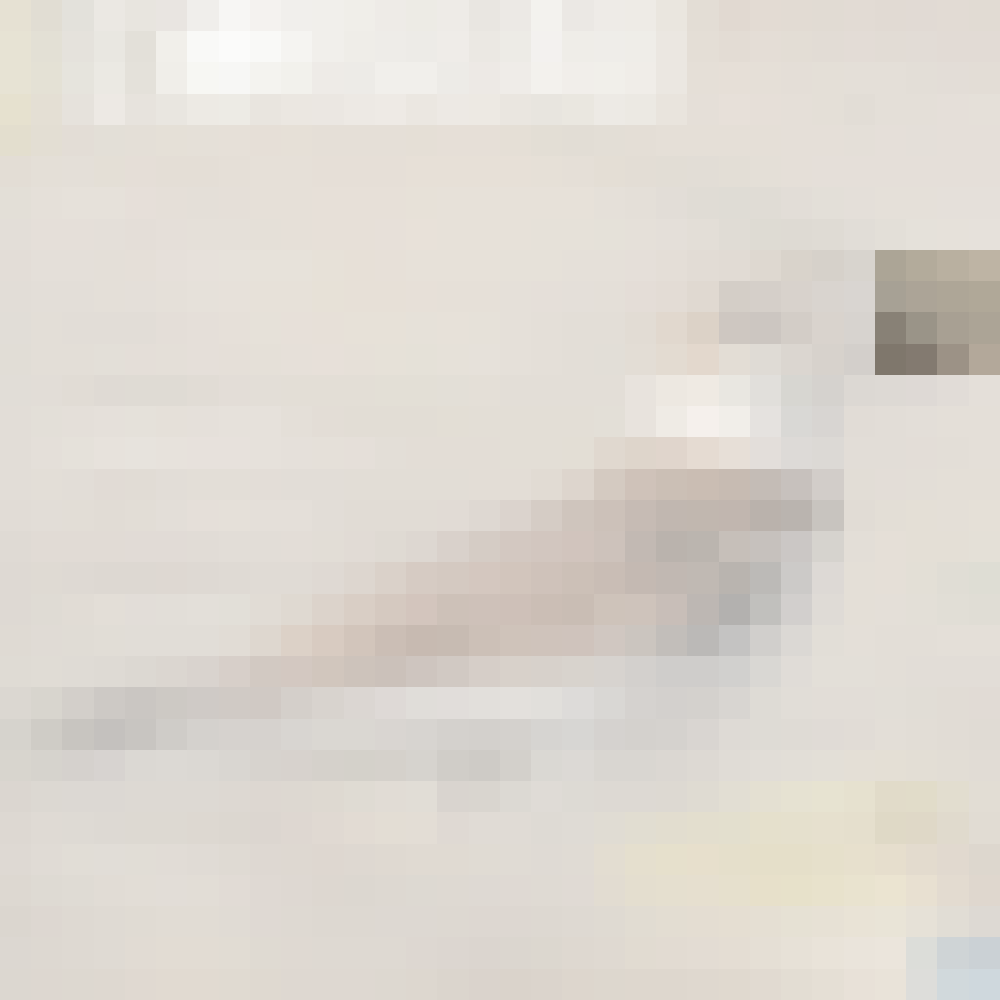}};
  \node at (2*1.525,  0*1.525) {\includegraphics[width=1.45cm]{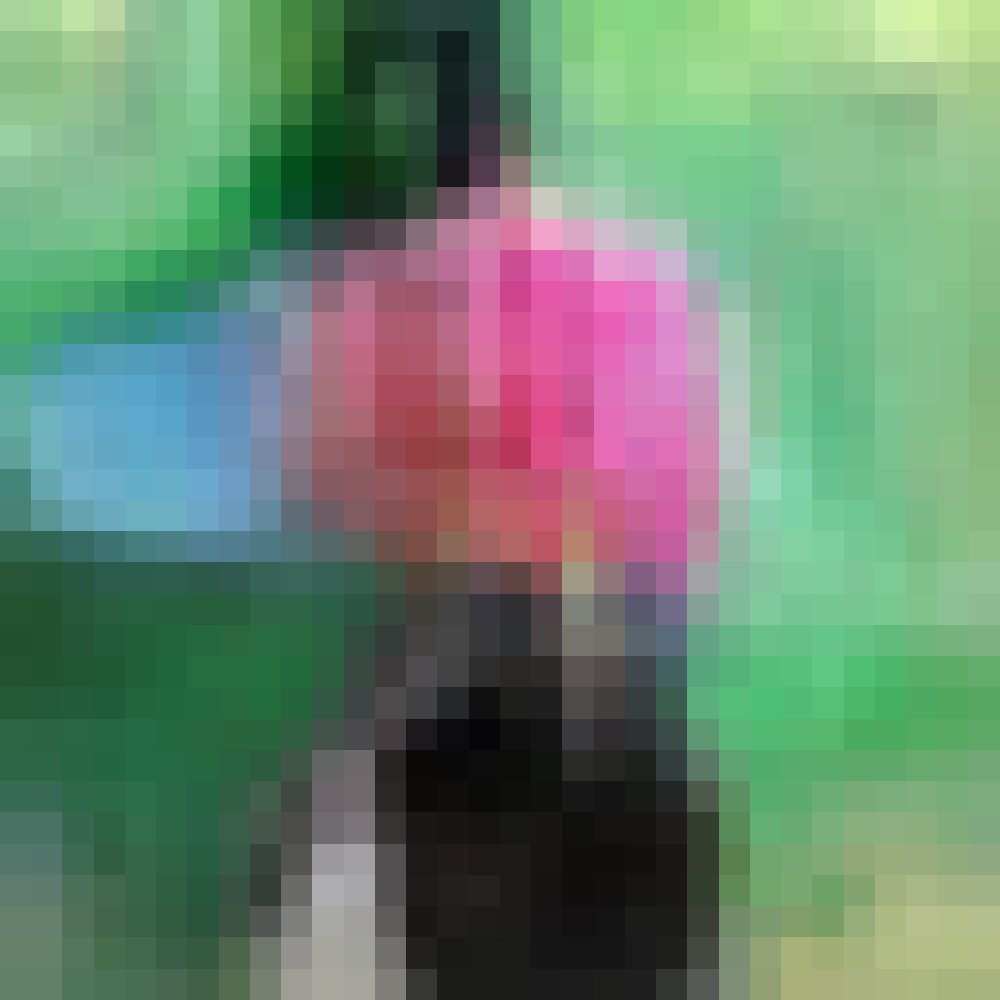}};
  \node at (3*1.525,  0*1.525) {\includegraphics[width=1.45cm]{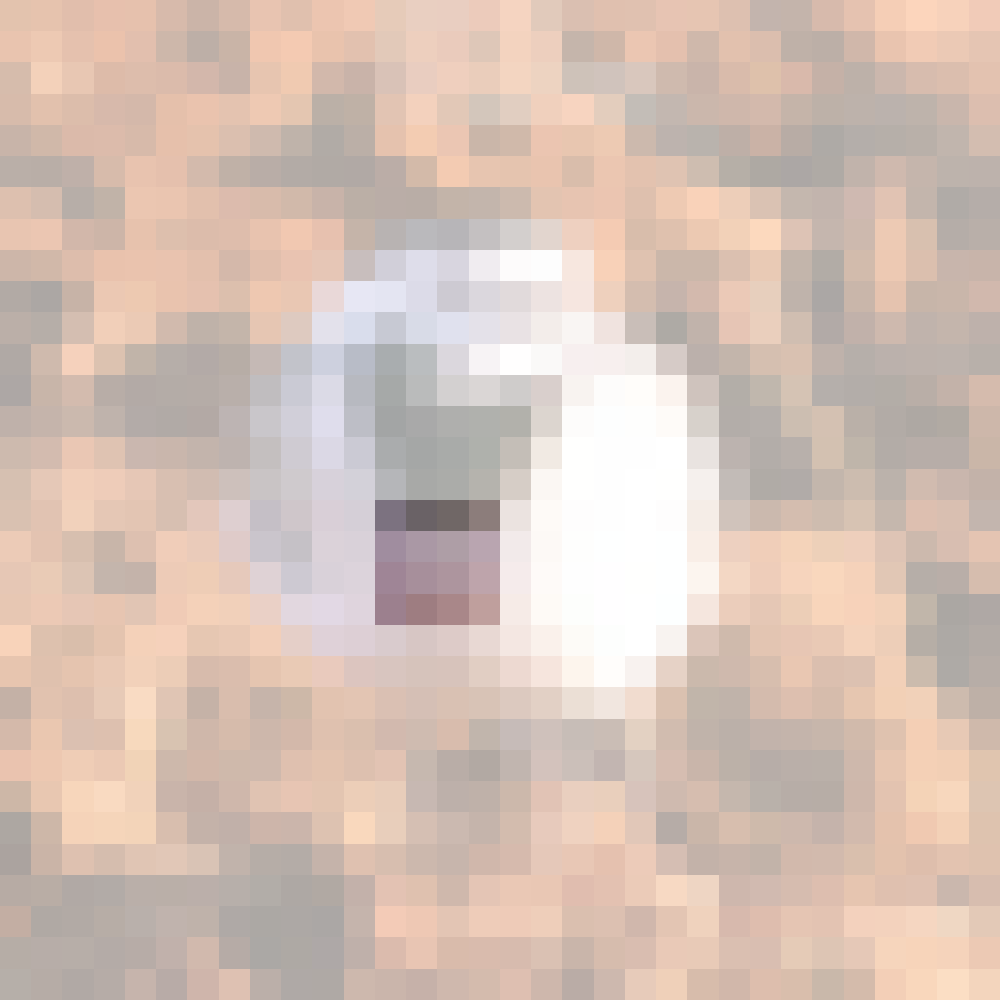}};
  \node at (0*1.525, -1*1.525) {\includegraphics[width=1.45cm]{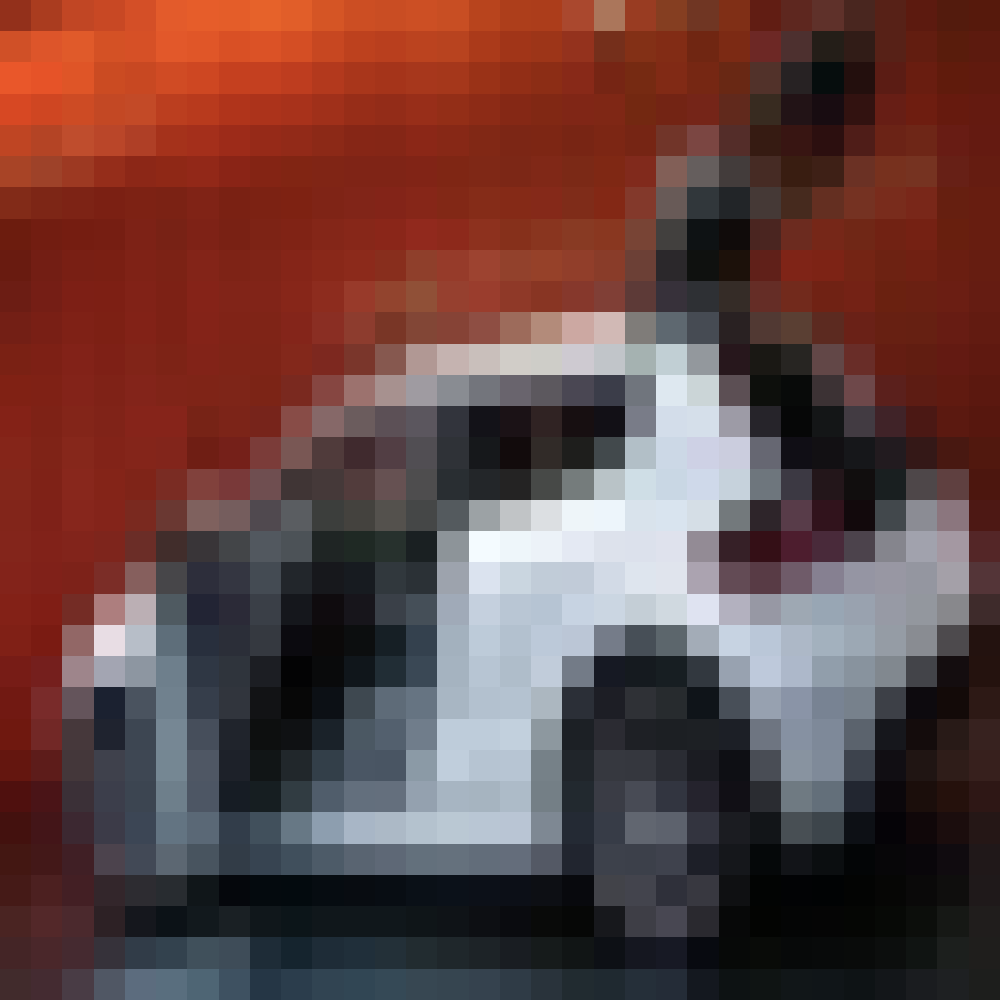}};
  \node at (1*1.525, -1*1.525) {\includegraphics[width=1.45cm]{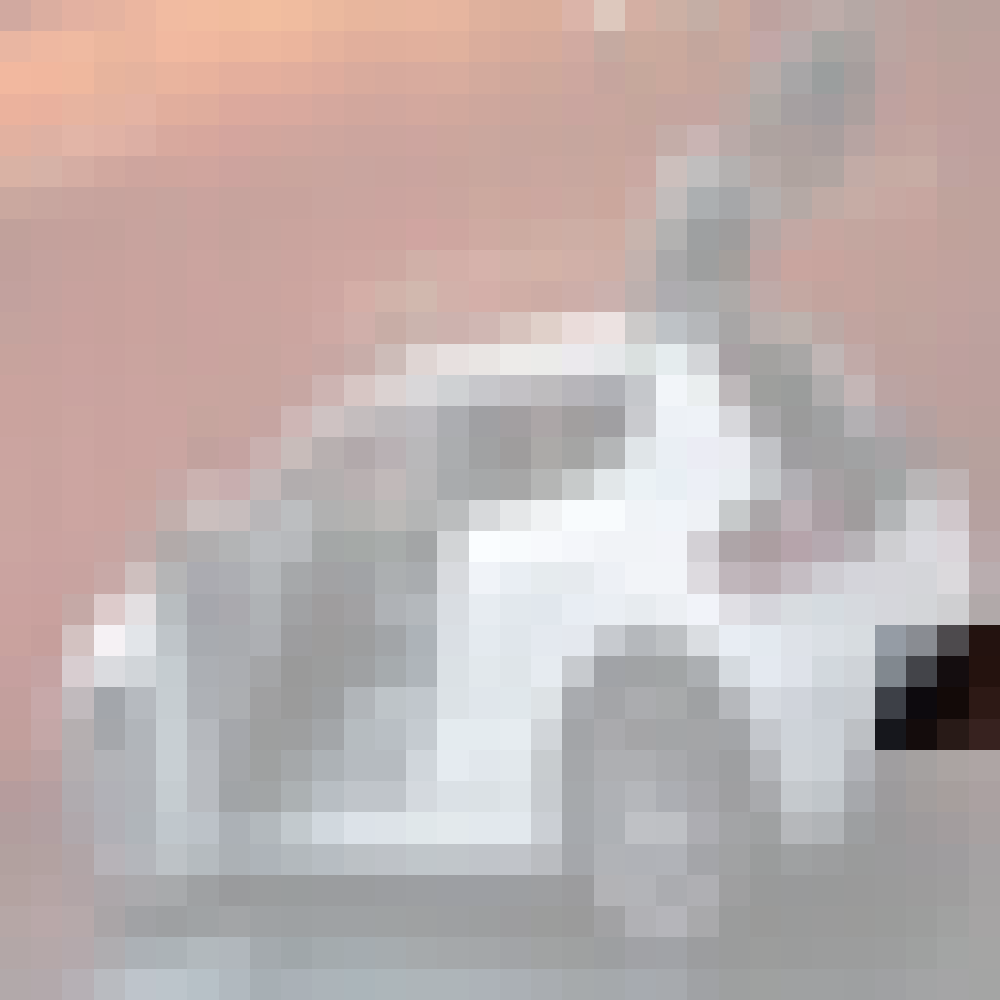}};
  \node at (2*1.525, -1*1.525) {\includegraphics[width=1.45cm]{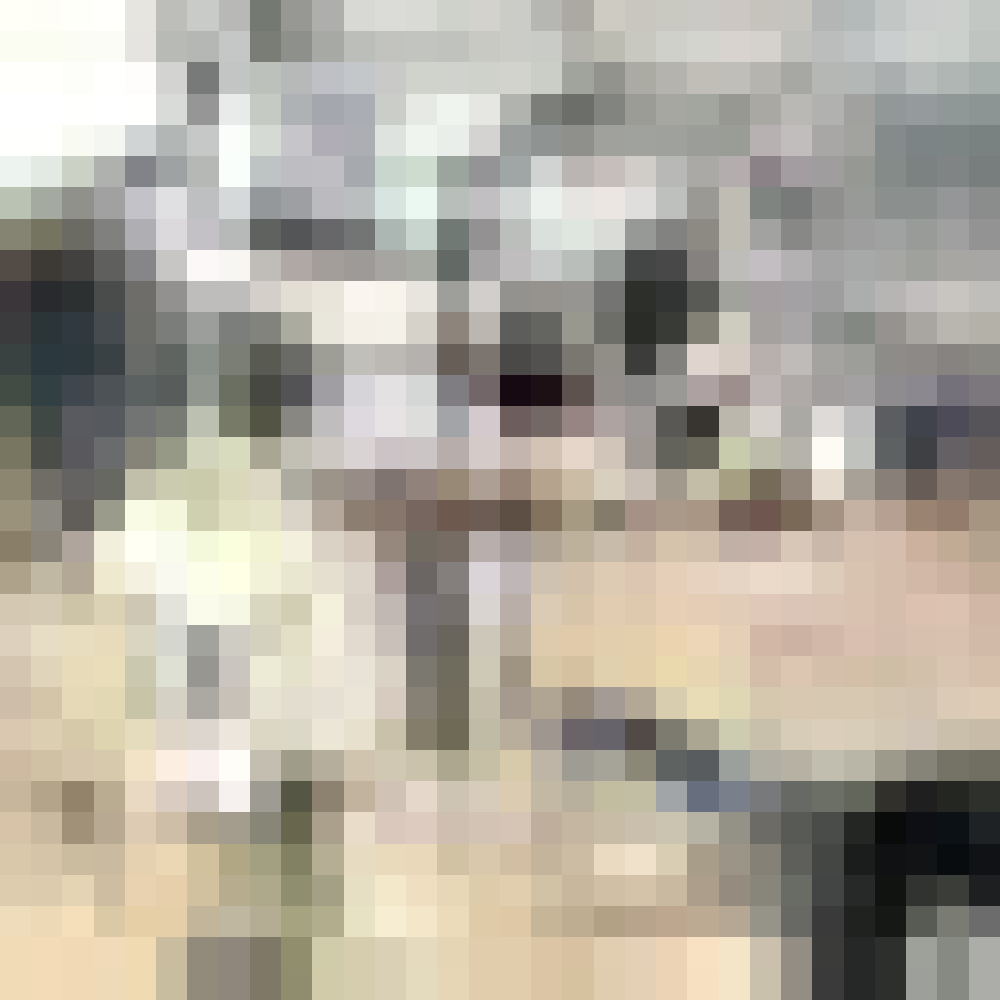}};
  \node at (3*1.525, -1*1.525) {\includegraphics[width=1.45cm]{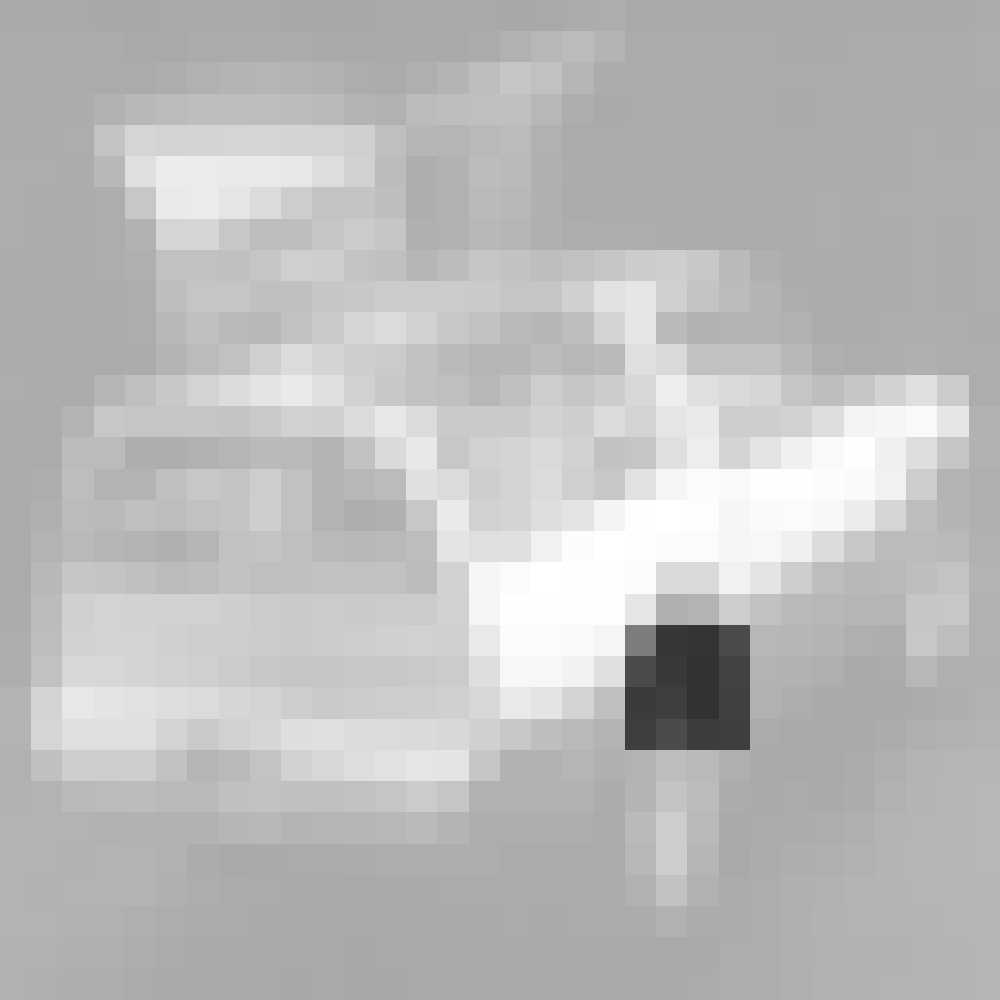}};

  \node at (-0.9,  0*1.525) {\tiny \strut \rotatebox{90}{\texttt{bird}}};
  \node at (-0.9, -1*1.525) {\tiny \strut \rotatebox{90}{\texttt{automobile}}};
  
  \node at (0*1.525, 0.9) {\tiny \strut Query\! Image};
  \node at (1*1.525, 0.9) {\tiny \strut Query\! Part};
  \node at (2*1.525, 0.9) {\tiny \strut Top\! Mean};
  \node at (3*1.525, 0.9) {\tiny \strut Retrieved\! Part};
\end{tikzpicture}

%% file: tikz/protobirb.tex
  \begin{tikzpicture}
    \node at (-0.5*1.525,  0*1.525) {\phantom{henlo}};
    \node at (-0.5*1.525, -1*1.525) {\phantom{henlo}};
    \node at ( 0*1.525,  0*1.525) {\includegraphics[width=1.45cm]{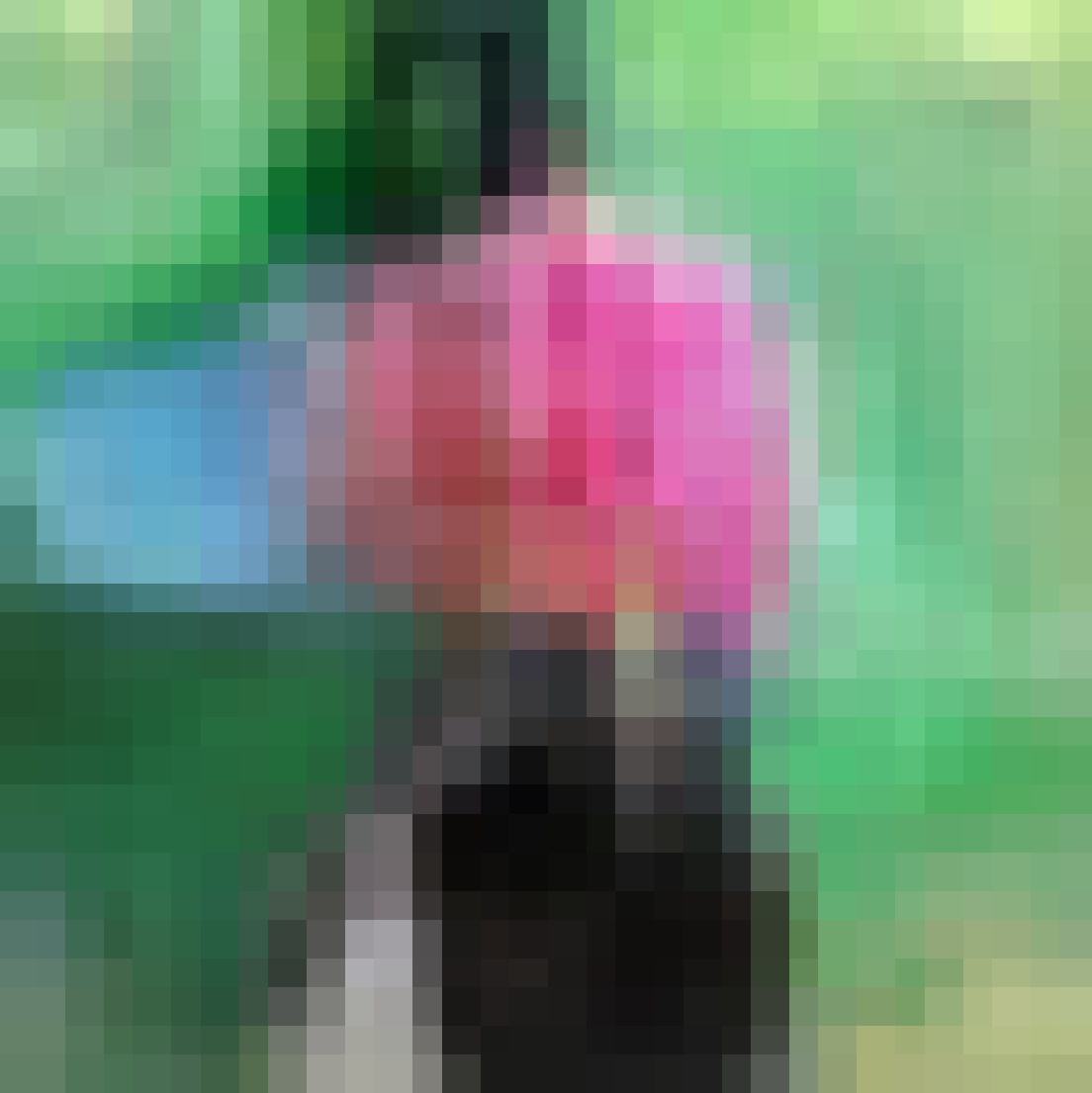}};
    \node at ( 0*1.525, -1*1.525) {\includegraphics[width=1.45cm]{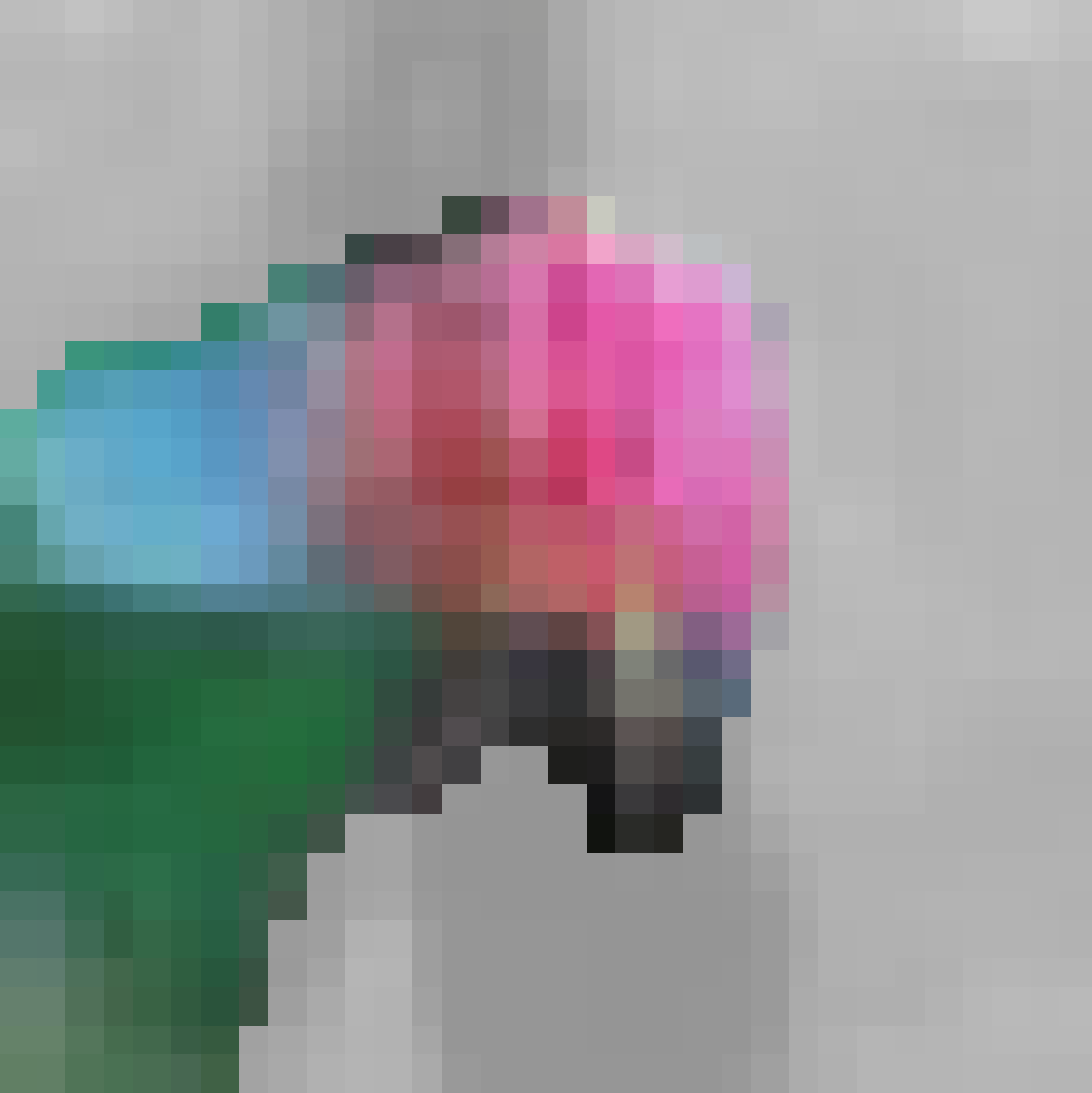}};
    \node at ( 0.5*1.525,  0*1.525) {\phantom{henlo}};
    \node at ( 0.5*1.525, -1*1.525) {\phantom{henlo}};
  \end{tikzpicture}

%% file: tikz/protobirbs.tex
  \begin{tikzpicture}
    \node at (0*1.525,  0*1.525) {\includegraphics[width=1.45cm]{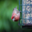}};
    \node at (0*1.525, -1*1.525) {\includegraphics[width=1.45cm]{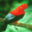}};
    \node at (1*1.525,  0*1.525) {\includegraphics[width=1.45cm]{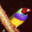}};
    \node at (1*1.525, -1*1.525) {\includegraphics[width=1.45cm]{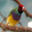}};
  \end{tikzpicture}

%% file: src/04.Experiments.tex
\section{Experiments and Analysis}
\label{sec:experiments}

\input{tables/clf_results}

We evaluate \Ours{} on a variety of image classification datasets used to evaluate prior joint predictive and generative modeling approaches: MNIST~\cite{deng2012mnist}, CIFAR-10~\cite{krizhevsky2009learning}, and CIFAR-100~\cite{krizhevsky2009learning}.
We also consider more challenging datasets, including Flowers-102~\cite{nilsback2008automated}, Oxford-IIIT Pet~\cite{parkhi2012cats}, and CUB-200-2011~\cite{WahCUB_200_2011}.

\paragraph{Setup}
We perform transfer learning by initializing the DenseFlow backbone with weights pre-trained for unconditional density estimation on $64 {\times} 64$ and $32 {\times} 32$ ImageNet~\cite{imagenet}.
Images are appropriately resized to these resolutions.
As DenseFlow is stochastic, we employ Monte Carlo sampling to reduce its variance.
We also use test-time augmentation.
All hyperparameters and other reproducibility details are provided in the supplemental material.
Code, configurations, and trained models are available at \url{https://github.com/craymichael/ProtoFlow}.

\paragraph{Predictive and Generative Performance}
We evaluate model performance using metrics that assess both predictive performance and generative performance.
Classification accuracy is used to quantify the former.
For the latter, we consider bits per dimension~(BPD), which is a common density estimation metric for normalizing flows.
BPD describes how many bits would be needed to encode a particular image in the modeled distribution~\cite{DBLP:journals/corr/TheisOB15}.
The results across all datasets are given in \cref{tab:clfResults}.
For existing generative classifiers based on normalizing flows, our approach achieves state-of-the-art accuracy and competitive BPD on all tasks.
We also establish baselines on additional challenging datasets.
\Ours{} outperforms other prototypical neural networks that report results on CIFAR-10.

\paragraph{Uncertainty and Calibration}
Reliable uncertainty quantification is critical in many applications, especially when algorithmic modeling leads to high-stakes decision-making.
In turn, we evaluate the reliability of predictive uncertainty by using the expected (ECE) and maximum (MCE) calibration 
These measures quantify the expected and maximum deviation of the predicted probabilities from the ground truth distribution, respectively.
As shown in \cref{tab:clfResults}, our approach has relatively low calibration error across all datasets.

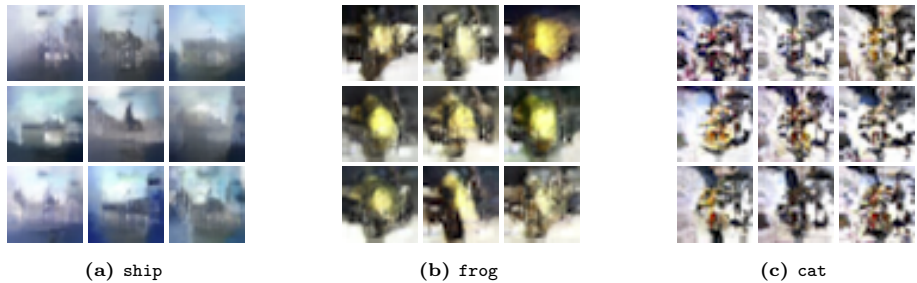
\begin{figure}[t]
  \captionsetup[subfigure]{aboveskip=-1.75ex}
  \centering
  \begin{subfigure}{.27\linewidth}
    \input{tikz/ship}
    \caption{\texttt{ship}}
    \label{fig:boat_vis_1}
  \end{subfigure}
  \hfill
  \begin{subfigure}{.27\linewidth}
    \input{tikz/frog}
    \caption{\texttt{frog}}
    \label{fig:frog_vis_1}
  \end{subfigure}
  \hfill
  \begin{subfigure}{.27\linewidth}
    \input{tikz/cat}
    \caption{\texttt{cat}}
    \label{fig:cat_vis_1}
  \end{subfigure}
  \caption{Mean points (center) and generated samples (periphery) from prototype distributions learned on CIFAR-10 with consistency loss and a truncation value of $1$.}
  \label{fig:proto_visualization_1}
\end{figure}

\input{tables/results}

\paragraph{Interpretability}
As discussed in \cref{subsec:gmm}, learned prototype distributions can be visualized faithfully and accurately by applying the inverse transform $\inv{f}$ to points sampled from the latent space.
We can influence the diversity and quality of these samples by drawing from truncated versions of the learned Gaussian distributions.
This technique---sometimes called the \emph{``truncation trick''}~\cite{brock2018large}---is often used when sampling from GANs.
Mean points and samples from prototypical distributions taken using this process are shown in \cref{fig:proto_visualization_1}.
For a more thorough analysis on the effects of truncation, please see the supplementary materials.

\begin{figure}[t]
  \begin{minipage}{0.48\textwidth}
    \centering
    \includegraphics[width=\linewidth]{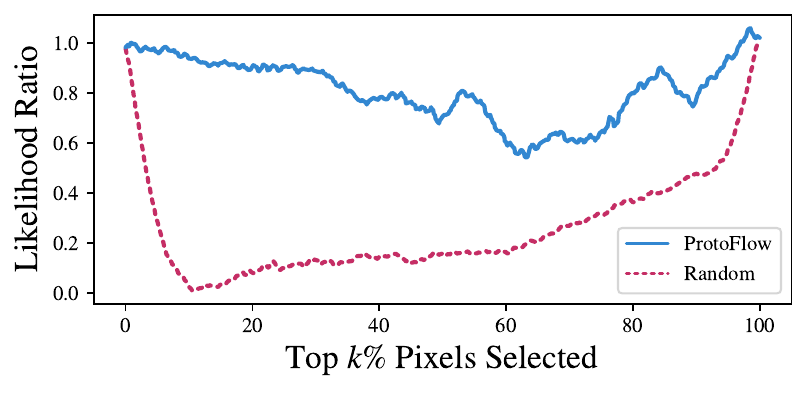}
    \caption{%
      Relevance ordering test for \Ours{} trained on CIFAR-10.
      Our heatmap approach discovers important image parts with respect to each prototypical distribution as substantiated by the likelihood gap with random heatmaps.
    }
    \label{fig:cifar-rot}
  \end{minipage}\hfill%
  \begin{minipage}{0.48\textwidth}
    \centering
    \includegraphics[width=\linewidth]{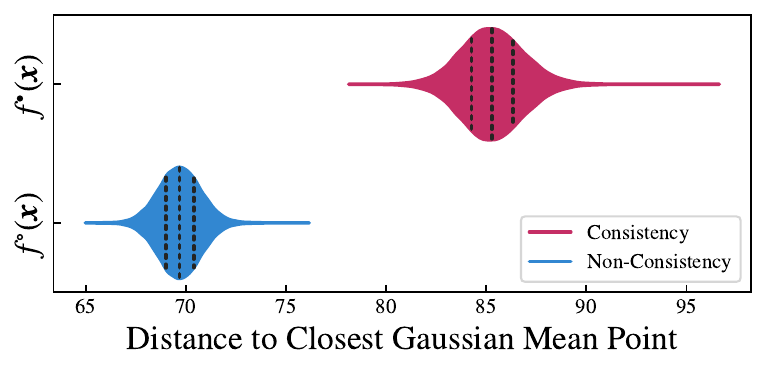}
    \caption{%
      Distances from each embedded point to its nearest prototypical mean point plotted for a model trained \emph{with} (in red) and \emph{without} (in blue) consistency regularization.
      Quartiles are shown as dashed lines.
    }
    \label{fig:distance_closest_proto}
  \end{minipage}
\end{figure}

On CIFAR-10, we explore our proposed prototypical parts approach.
See \cref{fig:cifar10-proto-parts} for examples of prototypical part explanations---the beak of a bird and the bumper of a car with high and low likelihood, respectively, are emphasized by \Ours{}.
The relevance ordering test is a quantitative measure of how well a heatmap attributes individual pixels according to prototype likelihoods~\cite{PRP}.
Starting from a completely random image, pixels are added back to the random image one at a time in descending order according to the heatmap $\mathcal{H}^k$.
As each pixel is added back, the likelihood of $\embedding$ conditioned on prototypical distribution $k$ is evaluated.
This procedure is averaged over each class-specific prototype over the data.
The faster the likelihood increases, the better the heatmap is.
\cref{fig:cifar-rot} shows the results of this test---our approach identifies important image parts substantially faster than random.
Interestingly, a uniformly random image has high likelihood, which lowers quickly as the pixels are added back randomly.
For a deeper understanding, please see the supplemental visualizations.

\paragraph{Robustness to Noise}
Are prototype distributions learned by \Ours{} robust to noise?
If so, then predicted prototypes should not be affected by small variations in the data.
Let $\noise \sim \mathcal{N}(0, s^2\mat{I})$ be multivariate Gaussian noise and $\random{x} \sim \pdata$ be random data.
The maximum likelihood prototype for $\random{x}$ is given by $\ML(\random{x})$ below.
\begin{equation}\label{eq:ml}
  \ML(\random{x})
  = \argmax_{(c, k) \in \classes \times \components} \platent\left(f(\random{x}) \mid \proto_{c, k} \right).
\end{equation}
We set the noise variance $s = 0.2$ for experiments, inspired by prior work~\cite{huang2023evaluation,Carmichael_2024_WACV}.
We define a robustness score, conceptually similar to a stability score~\cite{huang2023evaluation}, below.
\begin{equation}
  \score_{\rob}
  = \E_\proto \left\llbracket \ML(\random{x}) = \ML(\random{x} + \noise) \right\rrbracket
\end{equation}
The Iverson brackets $\llbracket\cdot\rrbracket$ above evaluate to $1$ or $0$ when the statement within the brackets is true or false, respectively.
$S_{\rob}$ represents the expected proportion of the data whose most likely prototype is stable under perturbation.
\cref{tab:scoreResults} shows \Ours{} is fairly robust, with $\score_{\rob} \gtrapprox 90\%$ across datasets.

\paragraph{Diversity}
We propose a new score to quantify the diversity of the learned prototypical distributions based on the Shannon entropy $H$ taken over prototypes.
\begin{equation}
  \score_{\div} = \frac{H\left( \ML(\random{x}) ; \proto \right)}{\log(KC)}
  = \frac{
    -\E_\proto
    \big[
      p\left( \ML\left(\random{x} \right) \right)
      \log{p\left( \ML(\random{x}) \right)}
    \big]
  }{\log(KC)}
\end{equation}
A lower score indicates redundant or uninformative prototypical distributions, and higher scores suggest diversity among prototypes.
In the worst case, there is a single prototypical distribution that maximizes the likelihood of every training sample.
In the best case, samples are evenly spread out across prototype distributions, yielding $H(\ML(\random{x}); \proto) = \log(KC)$.
We normalize by this maximum entropy to bound $\score_{\div}$ between $0$ and $1$.
\cref{tab:scoreResults} shows $\score_{\div}$ varies between datasets and is far from ideal, suggesting \Ours{} is learning redundant prototypes.

\paragraph{Prototype Pruning}
We evaluate the impact of prototype pruning on predictive and generative modeling performance.
We hypothesized that models might learn uninformative prototypes that could be removed without substantial performance degradation.
We confirm this hypothesis in \cref{tab:pruneResults}.
In particular, ${\sim}90\%$ of prototypes can be removed from a model trained on CIFAR-100 with ${<}0.2\%$ drop in accuracy.
This reinforces our prior result regarding prototype diversity.

\begin{figure}[t]
  \begingroup
    \captionsetup[subfigure]{aboveskip=-1.75ex, belowskip=1.75ex}
    \centering
    \begin{subfigure}{0.48\linewidth}
      \centering
      \input{tikz/consistency_mean_with}
      \caption{Means with consistency regularization.}
      \label{fig:consistency-means}
    \end{subfigure}%
    \hfill
    \begin{subfigure}{0.48\linewidth}
      \centering
      \input{tikz/consistency_sample_with}
      \caption{Samples with consistency regularization.}
      \label{fig:consistency-samples}
    \end{subfigure}
  \endgroup
  \begingroup
    \captionsetup[subfigure]{aboveskip=-1.75ex}
    \begin{subfigure}{0.48\linewidth}
      \centering
      \input{tikz/consistency_mean_without}
      \caption{Means without consistency regularization.}
      \label{fig:non-consistency-means}
    \end{subfigure}%
    \hfill%
    \begin{subfigure}{0.48\linewidth}
      \centering
      \input{tikz/consistency_sample_without}
      \caption{Samples without consistency regularization.}
      \label{fig:non-consistency-samples}
    \end{subfigure}
  \endgroup
  \caption{%
    A visual comparison of prototypical distributions on CIFAR-10 learned with and without consistency regularization.
    The consistency mean points \textbf{(a)} seem more interpretable than the uninformative means \textbf{(c)} learned without consistency.
    Despite this, samples with consistency \textbf{(b)} are poor compared to non-consistency samples \textbf{(d)}.
  }
\end{figure}
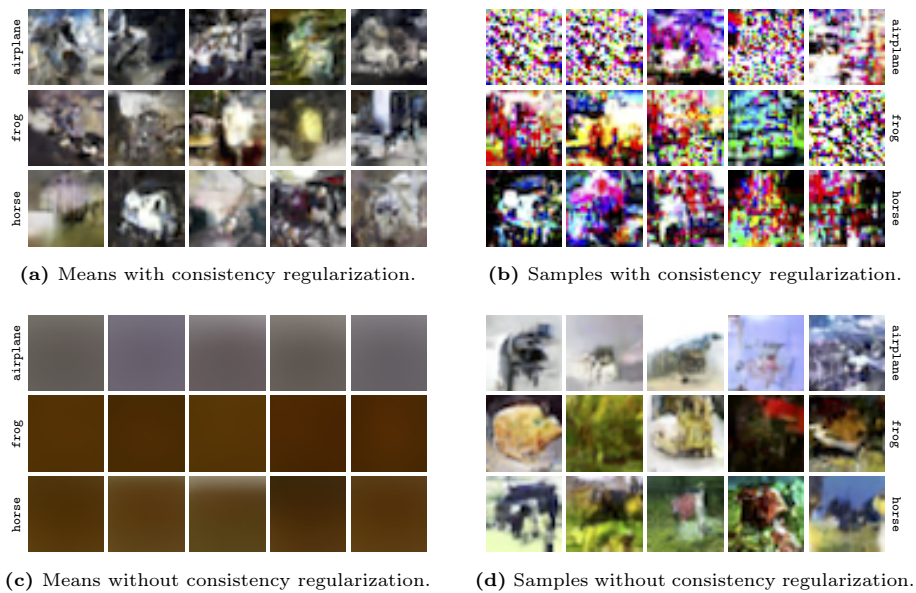

\paragraph{Impact of Consistency Loss}
To test the impact of the consistency regularization term $\loss_\con$, we compare a version of \Ours{} trained with consistency regularization $\wcon{f}$ to a version $\wocon{f}$ trained without it.
Both models are trained on CIFAR-10 with the same hyperparameters.
We visualize mean points and samples from a random subset of the corresponding prototypes and classes.
The visualizations of $\wcon{f}$ and $\wocon{f}$ are shown in \hyperref[fig:consistency-samples]{Fig.~6} and \hyperref[fig:non-consistency-samples]{Fig.~7}, respectively.
Immediately, we see an interesting pattern.
The prototypes learned \emph{with} $\loss_\con$ have good-looking mean point visualizations, but naively sampling from those distributions produces noise---this motivated our use of truncated sampling in models trained with $\loss_\con$ (\textit{cf.} \cref{fig:proto_visualization_1}).
Contrast this with the uninformative and blurry prototype means learned \emph{without} this loss term, which surprisingly yield high-quality samples that (1) correspond visually with their respective class and (2) vary predictably and sensibly between classes.
Since this behavior is consistent across prototypes within each model, we hypothesized that $\loss_\con$ affected the spatial distribution of embedded latent points relative to the learned prototypes.
For each data point $\data_i \in \domain$, we compute $\min_{\proto_{c, k}}\norm{\data_i - \mean_{c, k}}$ and plot these distributions in \cref{fig:distance_closest_proto}.
The model $\wocon{f}$ trained \emph{without} $\loss_\con$ tended to embed the data points significantly closer in $\latent$ to the means of prototypical distributions than $\wcon{f}$ did.
This helps explain why samples from $\wocon{f}$ appear more like real images.

This further highlights the flaw with point-based prototype explanations.
The noise in the samples visualized from $\wcon{f}$ suggests the regions of $\latent$ corresponding to those prototypes are not actually informative about the model---at least to a human.
Comparing these samples to their mean points reveals just how misleading a single point can be, possibly even signaling instances of overfitting.
Conversely, despite uninformative mean points, the samples taken from $\wocon{f}$ are consistently and undeniably more meaningful.

\paragraph{Additional Analyses and Results}
Please see the supplemental material for analyses of GMM initialization, results on CUB-200-2011, ablation of auxiliary loss terms, effect of Monte Carlo sampling, prototype visualizations, and more.

%% file: tables/clf_results.tex
\begin{table}[t]
    \ra{1.2}
    \tiny
    \centering
    \addtolength{\tabcolsep}{2pt}
    \begin{tabular}{@{}lllcccccc@{}}
        \toprule
        Dataset & Res & Model & Proto-Based & Flow-Based & Acc $\uparrow$ & BPD $\downarrow$ & ECE $\downarrow$ & MCE $\downarrow$ \\
        \midrule
        \multirow{5}{*}{\rotatebox{0}{MNIST}} & \multirow{5}{*}{$28\times 28$} & 
            \Ours{} (Ours) & \cmark & \cmark & {99.36} & \textbf{0.535} & 0.006 & 0.587 \\
          && FlowGMM~\cite{izmailov2020semi} & \xmark & \cmark & \textbf{99.63} & --- & \textbf{0.004}* & --- \\
          && Fetaya et al.~\cite{DBLP:conf/iclr/FetayaJGZ20} & \xmark & \cmark & 99.30 & {1.00} & --- & --- \\
          && SCNF-GLOW~\cite{atanov2019semi} & \xmark & \cmark & 88.44 & 1.15 & --- & --- \\
          && SCNF-GMM~\cite{atanov2019semi} & \xmark & \cmark & 83.10 & 1.14 & --- & --- \\[1.5ex]
        \multirow{10}{*}{\rotatebox{0}{CIFAR-10}} & \multirow{10}{*}{$32\times 32$} &
            \Ours{} (Ours) & \cmark & \cmark & \textbf{91.54} & 3.95 & 0.083 & \textbf{0.494} \\
          && IB-INN ($\gamma \rightarrow \infty$)~\cite{NEURIPS2020_593906af} & \xmark & \cmark & 91.28 & 17.3 & 0.81 & 13.9 \\
          && IB-INN ($\gamma=1$)~\cite{NEURIPS2020_593906af} & \xmark & \cmark & 89.73 & 5.25 & 0.54 & 3.25 \\
          && FlowGMM~\cite{izmailov2020semi} & \xmark & \cmark & 88.44 & --- & \textbf{0.038}* & --- \\
          && Fetaya et al.~\cite{DBLP:conf/iclr/FetayaJGZ20} & \xmark & \cmark & 84.00 & \textbf{3.53} & --- & --- \\[1ex]
          && KMEx~\cite{gautam2023prototypical} & \cmark & \xmark & 85.3 & --- & --- & --- \\
          && ProtoPNet~\cite{protopnet,gautam2023prototypical} & \cmark & \xmark & 84.9 & --- & --- & --- \\
          && FLINT~\cite{parekh2021framework} & \cmark & \xmark & 84.7 & --- & --- & --- \\
          && ProtoVAE~\cite{gautam2022protovae} & \cmark & \xmark & 84.6 & --- & --- & --- \\[1.5ex]
        \multirow{3}{*}{\rotatebox{0}{CIFAR-100}} & \multirow{3}{*}{$32\times 32$} &
            \Ours{} (Ours) & \cmark & \cmark & \textbf{69.80} & 5.03 & \textbf{0.292} & \textbf{0.637} \\
          && IB-INN ($\gamma \rightarrow \infty$)~\cite{NEURIPS2020_593906af} & \xmark & \cmark & 66.22 & 18.4 & 0.62 & 16.8 \\
          && IB-INN ($\gamma=1$)~\cite{NEURIPS2020_593906af} & \xmark & \cmark & 57.43 & \textbf{4.93} & 0.58 & 7.04 \\[1.5ex]
        Flowers-102 & $64\times 64$ & \Ours{} (Ours) & \cmark & \cmark & {59.80} & 13.5 & {0.141} & {0.295} \\[1.5ex]
        Oxford-IIIT Pet & $64\times 64$ & \Ours{} (Ours) & \cmark & \cmark & {53.58} & 4.89 & {0.459} & {0.736} \\
        \bottomrule \\
    \end{tabular}
    \caption{Results of normalizing-flow-based joint generative and predictive models across image classification tasks. Results reported from other prototypical neural networks are shown for CIFAR-10. \Ours{} achieves state-of-the-art accuracy while retaining highly competitive density estimation and calibration scores.
    *The result is reported with temperature-based scaling of GMM variances~\cite{izmailov2020semi}.%
    }
    \label{tab:clfResults}
    \addtolength{\tabcolsep}{-2pt}
\end{table}

%% file: tikz/ship.tex
\begin{tikzpicture}[ultra thick, line cap=round]
  \node at (0*1.075, 0*1.075) (s0) {\includegraphics[width=1cm]{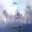}};
  \node at (1*1.075, 0*1.075) (s1) {\includegraphics[width=1cm]{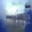}};
  \node at (2*1.075, 0*1.075) (s2) {\includegraphics[width=1cm]{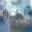}};
  \node at (0*1.075, 1*1.075) (s3) {\includegraphics[width=1cm]{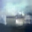}};
  \node at (1*1.075, 1*1.075) (s4) {\includegraphics[width=1cm]{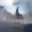}};
  \node at (2*1.075, 1*1.075) (s5) {\includegraphics[width=1cm]{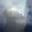}};
  \node at (0*1.075, 2*1.075) (s3) {\includegraphics[width=1cm]{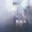}};
  \node at (1*1.075, 2*1.075) (s4) {\includegraphics[width=1cm]{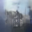}};
  \node at (2*1.075, 2*1.075) (s5) {\includegraphics[width=1cm]{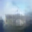}};
\end{tikzpicture}

%% file: tikz/frog.tex
\begin{tikzpicture}[ultra thick, line cap=round]
  \node at (0*1.075, 0*1.075) (s0) {\includegraphics[width=1cm]{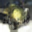}};
  \node at (1*1.075, 0*1.075) (s1) {\includegraphics[width=1cm]{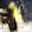}};
  \node at (2*1.075, 0*1.075) (s2) {\includegraphics[width=1cm]{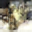}};
  \node at (0*1.075, 1*1.075) (s3) {\includegraphics[width=1cm]{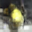}};
  \node at (1*1.075, 1*1.075) (s4) {\includegraphics[width=1cm]{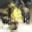}};
  \node at (2*1.075, 1*1.075) (s5) {\includegraphics[width=1cm]{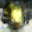}};
  \node at (0*1.075, 2*1.075) (s3) {\includegraphics[width=1cm]{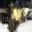}};
  \node at (1*1.075, 2*1.075) (s4) {\includegraphics[width=1cm]{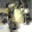}};
  \node at (2*1.075, 2*1.075) (s5) {\includegraphics[width=1cm]{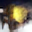}};
\end{tikzpicture}

%% file: tikz/cat.tex
\begin{tikzpicture}[ultra thick, line cap=round]
  \node at (0*1.075, 0*1.075) (s0) {\includegraphics[width=1cm]{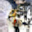}};
  \node at (1*1.075, 0*1.075) (s1) {\includegraphics[width=1cm]{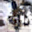}};
  \node at (2*1.075, 0*1.075) (s2) {\includegraphics[width=1cm]{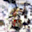}};
  \node at (0*1.075, 1*1.075) (s3) {\includegraphics[width=1cm]{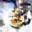}};
  \node at (1*1.075, 1*1.075) (s4) {\includegraphics[width=1cm]{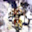}};
  \node at (2*1.075, 1*1.075) (s5) {\includegraphics[width=1cm]{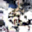}};
  \node at (0*1.075, 2*1.075) (s3) {\includegraphics[width=1cm]{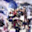}};
  \node at (1*1.075, 2*1.075) (s4) {\includegraphics[width=1cm]{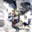}};
  \node at (2*1.075, 2*1.075) (s5) {\includegraphics[width=1cm]{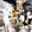}};
\end{tikzpicture}

%% file: tables/results.tex
\begin{table}[b]
  \parbox{.4\linewidth}{%
    \ra{1.2}
    \tiny
    \centering
    \addtolength{\tabcolsep}{4pt}
    \begin{tabular}[b]{@{}lcc@{}}
      \toprule
      Dataset & $\score_{\rob}$ $\uparrow$ & $\score_{\div}$ $\uparrow$ \\
      \midrule
      \multirow{1}{*}{{MNIST}}
      & 0.995 & 0.606 \\[1ex]
      \multirow{1}{*}{{CIFAR-10}}
      & 0.879 & 0.503 \\[1ex]
      \multirow{1}{*}{{CIFAR-100}}
      & 0.950 & 0.666 \\
      \bottomrule \\
    \end{tabular}
    \caption{\Ours{} robustness and diversity scores for learned prototypical distributions.}
    \label{tab:scoreResults}
    \addtolength{\tabcolsep}{-4pt}
  }%
  \hfill%
  \parbox{.55\linewidth}{%
    \centering
    \ra{1.2}
    \tiny
    \centering
    \addtolength{\tabcolsep}{4pt}
    \begin{tabular}[b]{@{}lccccc@{}}
      \toprule
      Dataset & \% Pruned & Acc $\uparrow$ & BPD $\downarrow$ & ECE $\downarrow$ & MCE $\downarrow$ \\
      \midrule
      \multirow{2}{*}{{MNIST}}
      & 0 & {99.39} & {0.535} & 0.006 & 0.587 \\
      & 54.0 & {96.26} & 1.13 & 0.025 & 0.591 \\[1ex]
      \multirow{2}{*}{{CIFAR-10}}
      & 0 & {91.54} & 3.95 & 0.083 & {0.494} \\
      & 78.0 & {91.13} & 3.95 & 0.087 & {0.614} \\[1ex]
      \multirow{2}{*}{{CIFAR-100}}
      & 0 & {69.80} & 5.03 & {0.292} & {0.637} \\
      & 89.9 & {69.63} & 5.02 & {0.294} & {0.637} \\
      \bottomrule \\
    \end{tabular}
    \caption{\Ours{} pruning results.}
    \label{tab:pruneResults}
    \addtolength{\tabcolsep}{-4pt}
  }
\end{table}

%% file: tikz/consistency_mean_with.tex
\begin{tikzpicture}[ultra thick, line cap=round]
  \node at (0*1.075, -0*1.075) (s0) {\includegraphics[width=1cm]{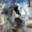}};
  \node at (1*1.075, -0*1.075) (s1) {\includegraphics[width=1cm]{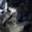}};
  \node at (2*1.075, -0*1.075) (s2) {\includegraphics[width=1cm]{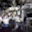}};
  \node at (3*1.075, -0*1.075) (s3) {\includegraphics[width=1cm]{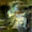}};
  \node at (4*1.075, -0*1.075) (s4) {\includegraphics[width=1cm]{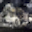}};
  
  \node at (0*1.075, -1*1.075) (s0) {\includegraphics[width=1cm]{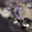}};
  \node at (1*1.075, -1*1.075) (s1) {\includegraphics[width=1cm]{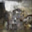}};
  \node at (2*1.075, -1*1.075) (s2) {\includegraphics[width=1cm]{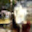}};
  \node at (3*1.075, -1*1.075) (s3) {\includegraphics[width=1cm]{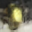}};
  \node at (4*1.075, -1*1.075) (s4) {\includegraphics[width=1cm]{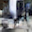}};
  
  \node at (0*1.075, -2*1.075) (s0) {\includegraphics[width=1cm]{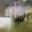}};
  \node at (1*1.075, -2*1.075) (s1) {\includegraphics[width=1cm]{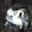}};
  \node at (2*1.075, -2*1.075) (s2) {\includegraphics[width=1cm]{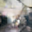}};
  \node at (3*1.075, -2*1.075) (s3) {\includegraphics[width=1cm]{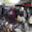}};
  \node at (4*1.075, -2*1.075) (s4) {\includegraphics[width=1cm]{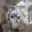}};

  \node[rotate=90] at (0*1.075 - 0.65, -0*1.075) {\tiny \strut \texttt{airplane}};
  \node[rotate=90] at (0*1.075 - 0.65, -1*1.075) {\tiny \strut \texttt{frog}};
  \node[rotate=90] at (0*1.075 - 0.65, -2*1.075) {\tiny \strut \texttt{horse}};
\end{tikzpicture}

%% file: tikz/consistency_sample_with.tex
\begin{tikzpicture}[ultra thick, line cap=round]
  \node at (0*1.075, -0*1.075) (s0) {\includegraphics[width=1cm]{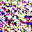}};
  \node at (1*1.075, -0*1.075) (s1) {\includegraphics[width=1cm]{figs/consistency_sample_with/prototype_samples_with1.png}};
  \node at (2*1.075, -0*1.075) (s2) {\includegraphics[width=1cm]{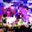}};
  \node at (3*1.075, -0*1.075) (s3) {\includegraphics[width=1cm]{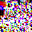}};
  \node at (4*1.075, -0*1.075) (s4) {\includegraphics[width=1cm]{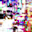}};
  
  \node at (0*1.075, -1*1.075) (s0) {\includegraphics[width=1cm]{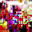}};
  \node at (1*1.075, -1*1.075) (s1) {\includegraphics[width=1cm]{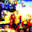}};
  \node at (2*1.075, -1*1.075) (s2) {\includegraphics[width=1cm]{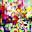}};
  \node at (3*1.075, -1*1.075) (s3) {\includegraphics[width=1cm]{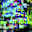}};
  \node at (4*1.075, -1*1.075) (s4) {\includegraphics[width=1cm]{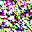}};
  
  \node at (0*1.075, -2*1.075) (s0) {\includegraphics[width=1cm]{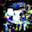}};
  \node at (1*1.075, -2*1.075) (s1) {\includegraphics[width=1cm]{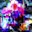}};
  \node at (2*1.075, -2*1.075) (s2) {\includegraphics[width=1cm]{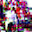}};
  \node at (3*1.075, -2*1.075) (s3) {\includegraphics[width=1cm]{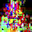}};
  \node at (4*1.075, -2*1.075) (s4) {\includegraphics[width=1cm]{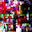}};

  \node[rotate=270] at (4*1.075 + 0.65, -0*1.075) {\tiny \strut \texttt{airplane}};
  \node[rotate=270] at (4*1.075 + 0.65, -1*1.075) {\tiny \strut \texttt{frog}};
  \node[rotate=270] at (4*1.075 + 0.65, -2*1.075) {\tiny \strut \texttt{horse}};
\end{tikzpicture}

%% file: tikz/consistency_mean_without.tex
\begin{tikzpicture}[ultra thick, line cap=round]
  \node at (0*1.075, -0*1.075) (s0) {\includegraphics[width=1cm]{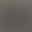}};
  \node at (1*1.075, -0*1.075) (s1) {\includegraphics[width=1cm]{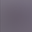}};
  \node at (2*1.075, -0*1.075) (s2) {\includegraphics[width=1cm]{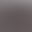}};
  \node at (3*1.075, -0*1.075) (s3) {\includegraphics[width=1cm]{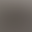}};
  \node at (4*1.075, -0*1.075) (s4) {\includegraphics[width=1cm]{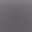}};
  
  \node at (0*1.075, -1*1.075) (s0) {\includegraphics[width=1cm]{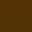}};
  \node at (1*1.075, -1*1.075) (s1) {\includegraphics[width=1cm]{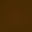}};
  \node at (2*1.075, -1*1.075) (s2) {\includegraphics[width=1cm]{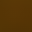}};
  \node at (3*1.075, -1*1.075) (s3) {\includegraphics[width=1cm]{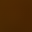}};
  \node at (4*1.075, -1*1.075) (s4) {\includegraphics[width=1cm]{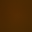}};
  
  \node at (0*1.075, -2*1.075) (s0) {\includegraphics[width=1cm]{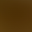}};
  \node at (1*1.075, -2*1.075) (s1) {\includegraphics[width=1cm]{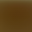}};
  \node at (2*1.075, -2*1.075) (s2) {\includegraphics[width=1cm]{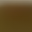}};
  \node at (3*1.075, -2*1.075) (s3) {\includegraphics[width=1cm]{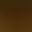}};
  \node at (4*1.075, -2*1.075) (s4) {\includegraphics[width=1cm]{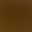}};

  \node[rotate=90] at (0*1.075 - 0.65, -0*1.075) {\tiny \strut \texttt{airplane}};
  \node[rotate=90] at (0*1.075 - 0.65, -1*1.075) {\tiny \strut \texttt{frog}};
  \node[rotate=90] at (0*1.075 - 0.65, -2*1.075) {\tiny \strut \texttt{horse}};
\end{tikzpicture}

%% file: tikz/consistency_sample_without.tex
\begin{tikzpicture}[ultra thick, line cap=round]
  \node at (0*1.075, -0*1.075) (s0) {\includegraphics[width=1cm]{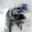}};
  \node at (1*1.075, -0*1.075) (s1) {\includegraphics[width=1cm]{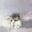}};
  \node at (2*1.075, -0*1.075) (s2) {\includegraphics[width=1cm]{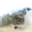}};
  \node at (3*1.075, -0*1.075) (s3) {\includegraphics[width=1cm]{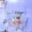}};
  \node at (4*1.075, -0*1.075) (s4) {\includegraphics[width=1cm]{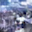}};
  
  \node at (0*1.075, -1*1.075) (s0) {\includegraphics[width=1cm]{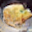}};
  \node at (1*1.075, -1*1.075) (s1) {\includegraphics[width=1cm]{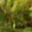}};
  \node at (2*1.075, -1*1.075) (s2) {\includegraphics[width=1cm]{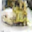}};
  \node at (3*1.075, -1*1.075) (s3) {\includegraphics[width=1cm]{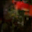}};
  \node at (4*1.075, -1*1.075) (s4) {\includegraphics[width=1cm]{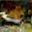}};
  
  \node at (0*1.075, -2*1.075) (s0) {\includegraphics[width=1cm]{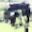}};
  \node at (1*1.075, -2*1.075) (s1) {\includegraphics[width=1cm]{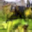}};
  \node at (2*1.075, -2*1.075) (s2) {\includegraphics[width=1cm]{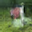}};
  \node at (3*1.075, -2*1.075) (s3) {\includegraphics[width=1cm]{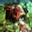}};
  \node at (4*1.075, -2*1.075) (s4) {\includegraphics[width=1cm]{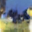}};

  \node[rotate=270] at (4*1.075 + 0.65, -0*1.075) {\tiny \strut \texttt{airplane}};
  \node[rotate=270] at (4*1.075 + 0.65, -1*1.075) {\tiny \strut \texttt{frog}};
  \node[rotate=270] at (4*1.075 + 0.65, -2*1.075) {\tiny \strut \texttt{horse}};
\end{tikzpicture}

%% file: src/05.Discussion.tex
\section{Discussion}
\label{sec:discussion}

Our proposed prototypical neural network \Ours{} achieves state-of-the-art performance on joint generative predictive modeling tasks.
By learning prototypes as latent-space probability distributions using normalizing flows, it simultaneously delivers rich, faithful visualizations and precise uncertainty quantification. 
We also establish baselines on more complex datasets than prior work\footnote{%
  IB-INN~\cite{mackowiak2021generative} was evaluated on ImageNet, but these results may not be reproducible.
  See the supplemental material for more detail and further discussion.
}.

A limitation of our implementation is the $64 \times 64$ pixel maximum on input image resolution, an artifact of the pre-trained DenseFlow models available.
We also fail to fully exploit invertibility during backpropagation---intermediate activations can be recomputed during the backward pass instead of being stored during the forward pass~\cite{behrmann2019invertible,mangalam2022reversible,etmann2020iunets,DBLP:conf/iclr/JacobsenSO18}.
Since \Ours{} is vulnerable to training data biases, we recommend appropriate safety and fairness precautions~\cite{10.1145/3457607,achiam2023gpt,amodei2016concrete}.

Future work should find it straight-forward to extend \Ours{} to semi-supervised settings by optimizing the joint likelihood over labeled and unlabeled data.
The number of prototypical distributions should be selected in a more principled way---\eg{}, using information theory~\cite{sugar2003finding,still2004many}.
Even by applying techniques to avoid overfitting, such as dropout, data augmentation, and weight regularization, it proved difficult to avoid a generalization gap on the Oxford-IIIT Pets and Flowers-102 datasets.
Subsequent research should consider more aggressive data augmentation and increasing the resolution of images used as input to \Ours{}. We are hopeful that this new avenue for prototype learning in neural networks will lead to further innovations that open the black box and close the human-machine semantic similarity gap.

%% file: src/99.Supplemental.tex
\appendix

\begin{table}
  \parbox{0.5\linewidth}{
    \centering
    \begin{tabular}{@{}lcr@{}}
      Hyperparameter            && Value   \\
      \toprule
      Learning Rate (LR)        && 0.0001  \\
      LR Schedule               && Cosine  \\
      Batch Size                && 256     \\
      Optimizer                 && AdamW   \\
      $K$                       && 10      \\
      Warmup Epochs             && 5       \\
      Maximum Grad Norm         && 100     \\
      Weight Decay              && 0.00001 \\
      Consistency Rampup        && Linear  \\
      Consistency Rampup Epochs && 50      \\
      \bottomrule                          \\
    \end{tabular}
    \caption{Experiment hyperparameters.}
    \label{tab:hparams}
  }\hfill%
  \parbox{0.5\linewidth}{
    \centering
    \begin{tabular}{@{}lcrcr@{}}
      $\beta$        && \textsc{Acc} Reported && \textsc{Acc} Reproduced \\
      \toprule
      1              && 67.30\%               && 25.30\%                 \\
      2              && 71.73\%               && 0.56\%                  \\
      4              && 73.69\%               && 0.21\%                  \\
      8              && 74.59\%               && 0.12\%                  \\
      16             && 75.54\%               && 0.76\%                  \\
      32             && 76.18\%               && 36.81\%                 \\
      $\infty$       && 76.27\%               && 74.40\%                 \\
      \bottomrule                                                        \\
    \end{tabular}
    \caption{IB-INN ImageNet discrepancy.}
    \label{tab:ibinn}
  }
\end{table}

\section{Reproducibility}

\Ours{} is trained on machines with up to four \textsc{Nvidia} A10 GPUs in parallel and an Intel Xeon Gold 6326 CPU.
We employ the following data augmentation scheme to each image during training.
\begin{enumerate}
  \item
    Resize the image's shortest axis to $32$ or $64$ pixels using bicubic interpolation.
  \item
    Apply a horizontal reflection randomly with probability $\sfrac{1}{2}$.
  \item
    Pad the image appropriately by duplicating pixels along the edges.
  \item
    Randomly perform one or more of the following affine transformations:  
    \begin{itemize}
      \item
        a rotation of $\theta^\circ$ degrees, where $\theta^\circ \in [-15^{\circ}, 15^{\circ}]$
      \item
        a horizontal/vertical translation by $t\%$ pixels, with $t\% \in [-4\%, 4\%]$
      \item
        a shear transformation (transvection) of $\alpha^\circ$ degrees, with $\alpha^\circ \in [-10^{\circ}, 10^{\circ}]$
    \end{itemize}
  \item
    Center-crop the image to $32\times32$ or $64\times64$ resolution.
\end{enumerate}
The DenseFlow architecture was instantiated with a block configuration of $[6, 4,1]$, layer middle channels of $[48,48,48]$, and a layer configuration of $[5,6,20]$.
During training, we use a single Monte Carlo sample in the forward pass of DenseFlow.
During testing, we report results with 100 Monte Carlo samples.
Furthermore, we use test-time augmentation, using the same augmentation composition as during training.
For each sample, we average logits and likelihood estimates across five augmentations.
All models were trained for 150 epochs unless otherwise specified.
For prototypical parts, we use a patch size of $4\times4$ pixels.
All other hyperparameters are listed in \cref{tab:hparams}.

\subsection{IB-INN Reproducibility Concerns}
IB-INN was evaluated on ImageNet~\cite{mackowiak2021generative}, which is a big step beyond the typical evaluations on MNIST and CIFAR-10.
However, we were unable to reproduce these results and the GitHub issue\footnote{\url{https://github.com/RayDeeA/ibinn_imagenet/issues/7}} about reproducibility has not been remedied since being opened in 2021.  %
\cref{tab:ibinn} indicates the reported model performance and the reproduced model performance.
Recall that $\beta$ is a loss weight hyperparameter that controls the trade-off between generative (lower values) and predictive (higher values) quality.

Per a suggestion in the aforementioned issue, we continued trained IB-INN ($\beta = 8$) for an additional 10 epochs using their provided training script.
However, the accuracy did not improve above $1\%$.
We did not try fine-tuning IB-INN with other values of $\beta$ due to the heavy training cost.
Since the only model with impressive accuracy has $\beta \rightarrow \infty$, which destroys the generative capability of the model, we did not use IB-INN as a backbone for \Ours{}.
Moreover, the authors of~\cite{peters2022extending} who employ IB-INN were unable to reproduce the reported accuracy scores.
Upon interpolating between samples in latent space, the authors find that the interpolations are very noisy and not apt for meaningful interpretation.

\section{Additional Results and Visualizations}
We compare the performance of our model against ProtoPNet on CUB-200-2011 in \cref{tab:ppResults}.
For both models, the bird bounding box annotations are used to crop the images.
We obtain the result for ProtoPNet by training it using the author-provided code base with a ResNet-34 backbone for 50 epochs.
With \Ours{}, we first fit DenseFlow to the CUB-200-2011 training set before training it for joint generation and image classification.
While our obtained accuracy is lower, our model offers richer prototype interpretation.
There are many considerations for future work to improve model performance and reduce overfitting, as discussed in the main text.

\input{tables/pp_results}

\input{tables/clf_results_full}

We also include a full results table to compare the generative classifiers to state-of-the-art prediction-only models.
This helps highlight how these two classes of models compare and the impact of generative and prototypical modeling aspects.
These results are shown in \cref{tab:clfResultsFull}.

\begin{figure}
  \centering
  \includegraphics[width=\linewidth]{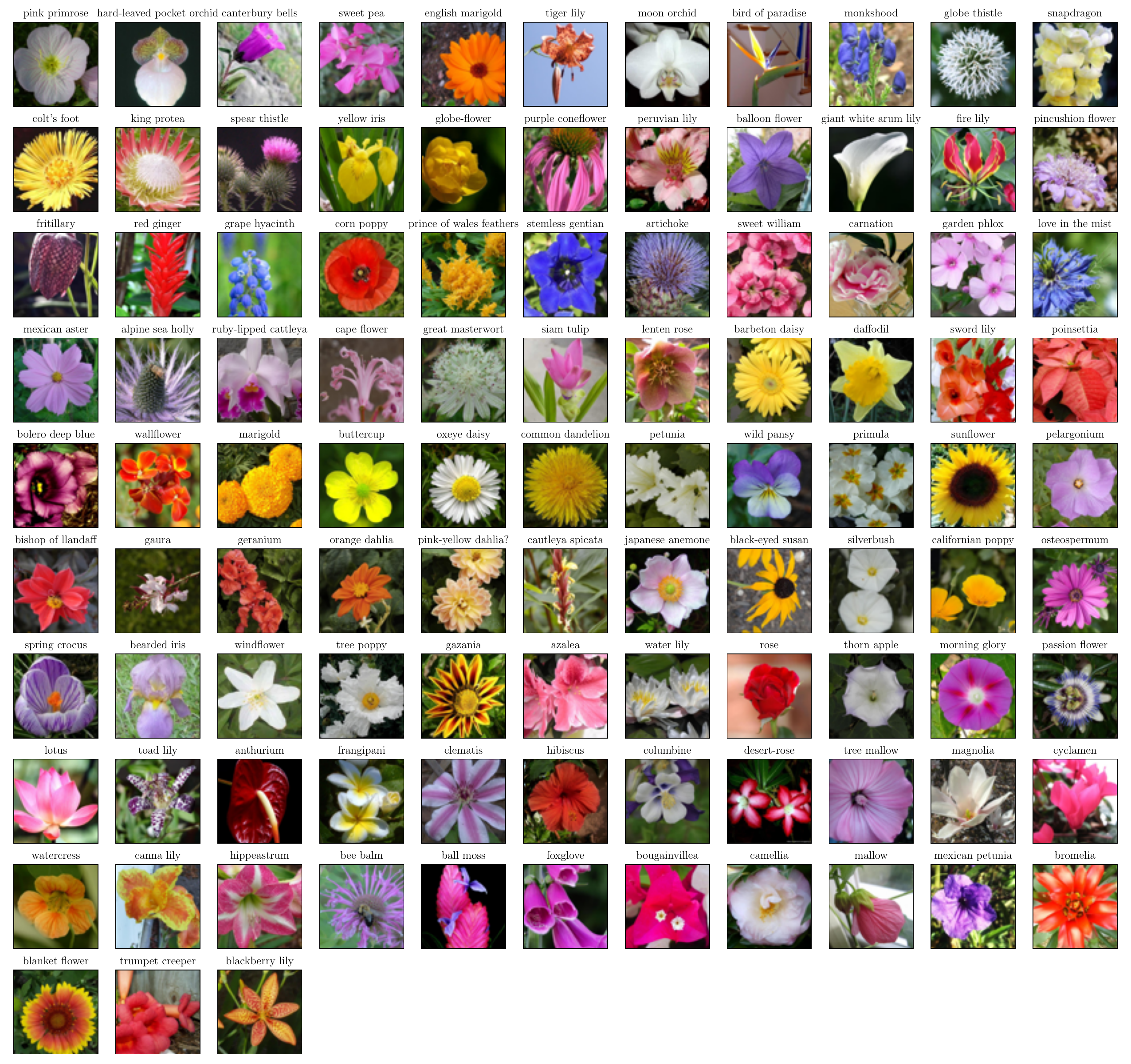}
  \caption{Most likely images from Flowers-102 for each class-specific GMM learned.}
  \label{fig:likely_samples_flowers}
\end{figure}

\begin{figure}
  \centering
  \begin{subfigure}{\linewidth}
    \centering
    \includegraphics[width=\linewidth]{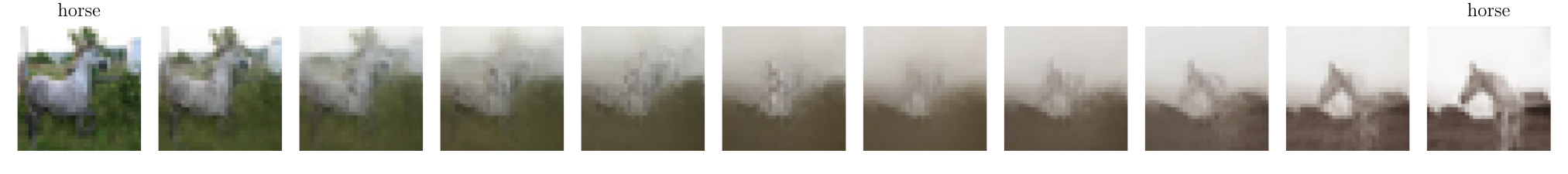}
    \caption{\texttt{horse} $\longleftrightarrow$ \texttt{horse}}
    \label{fig:interp_horse}
  \end{subfigure}
  \\
  \begin{subfigure}{\linewidth}
    \centering
    \includegraphics[width=\linewidth]{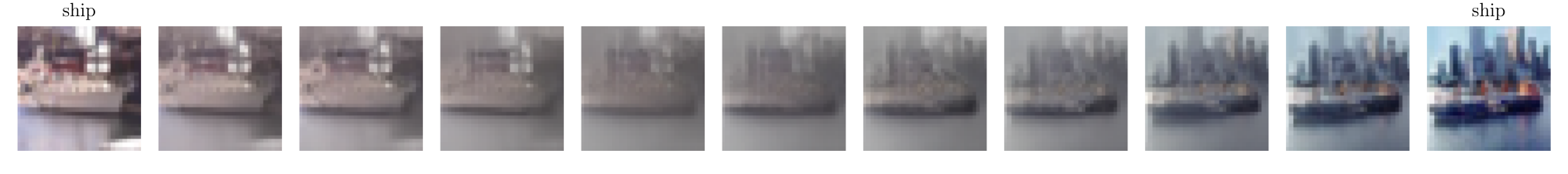}
    \caption{\texttt{ship} $\longleftrightarrow$ \texttt{ship}}
    \label{fig:interp_ship}
  \end{subfigure}
  \\
  \begin{subfigure}{\linewidth}
    \centering
    \includegraphics[width=\linewidth]{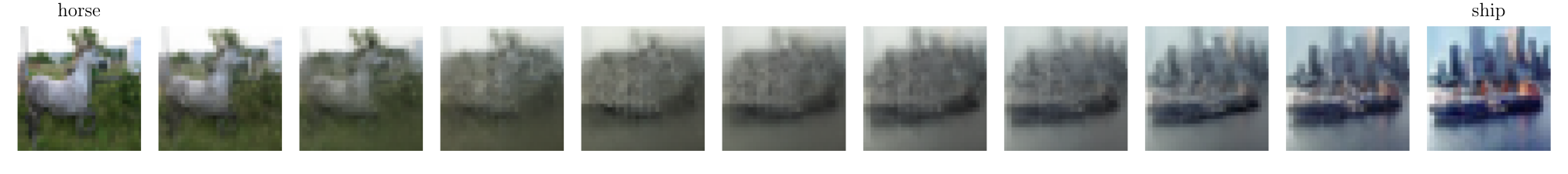}
    \caption{\texttt{horse} $\longleftrightarrow$ \texttt{ship}}
    \label{fig:interp_horse_ship}
  \end{subfigure}
  \\
  \begin{subfigure}{\linewidth}
    \centering
    \includegraphics[width=\linewidth]{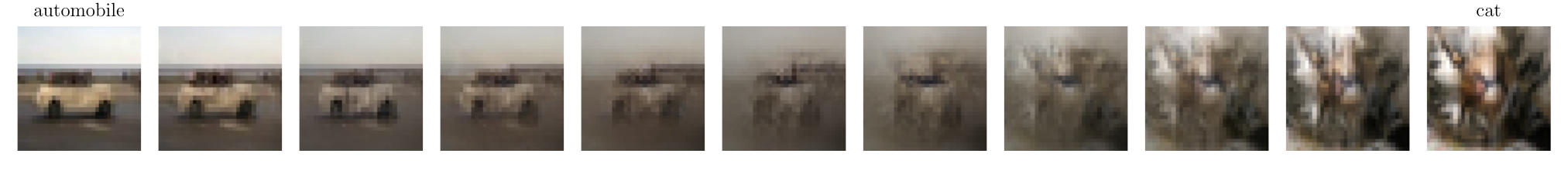}
    \caption{\texttt{automobile} $\longleftrightarrow$ \texttt{cat}}
    \label{fig:interp_car_cat}
  \end{subfigure}
  \caption{%
    Linear latent interpolations between pairs of images.
    Both intra-class and inter-class interpolations are shown for \Ours{} trained on CIFAR-10.
  }
  \label{fig:interpolate-viz}
\end{figure}

In addition, we visualize the most likely data samples according to each GMM in a trained model.
This helps to qualitatively understand what the most representative exemplar looks like according to the learned prototypical distributions.
\cref{fig:likely_samples_flowers} shows these samples for \Ours{} trained on Flowers-102.
Next, we visualize linear interpolations in the latent space between pairs of images with a random subset shown in \cref{fig:interpolate-viz}.
In all pairs, there is a smooth transition of both the foreground and background.
This also helps to understand the boundaries between classes, which, qualitatively, retain structural information of objects but few class-discriminative details.

\begin{figure}
    \centering
    \includegraphics[width=0.8\linewidth]{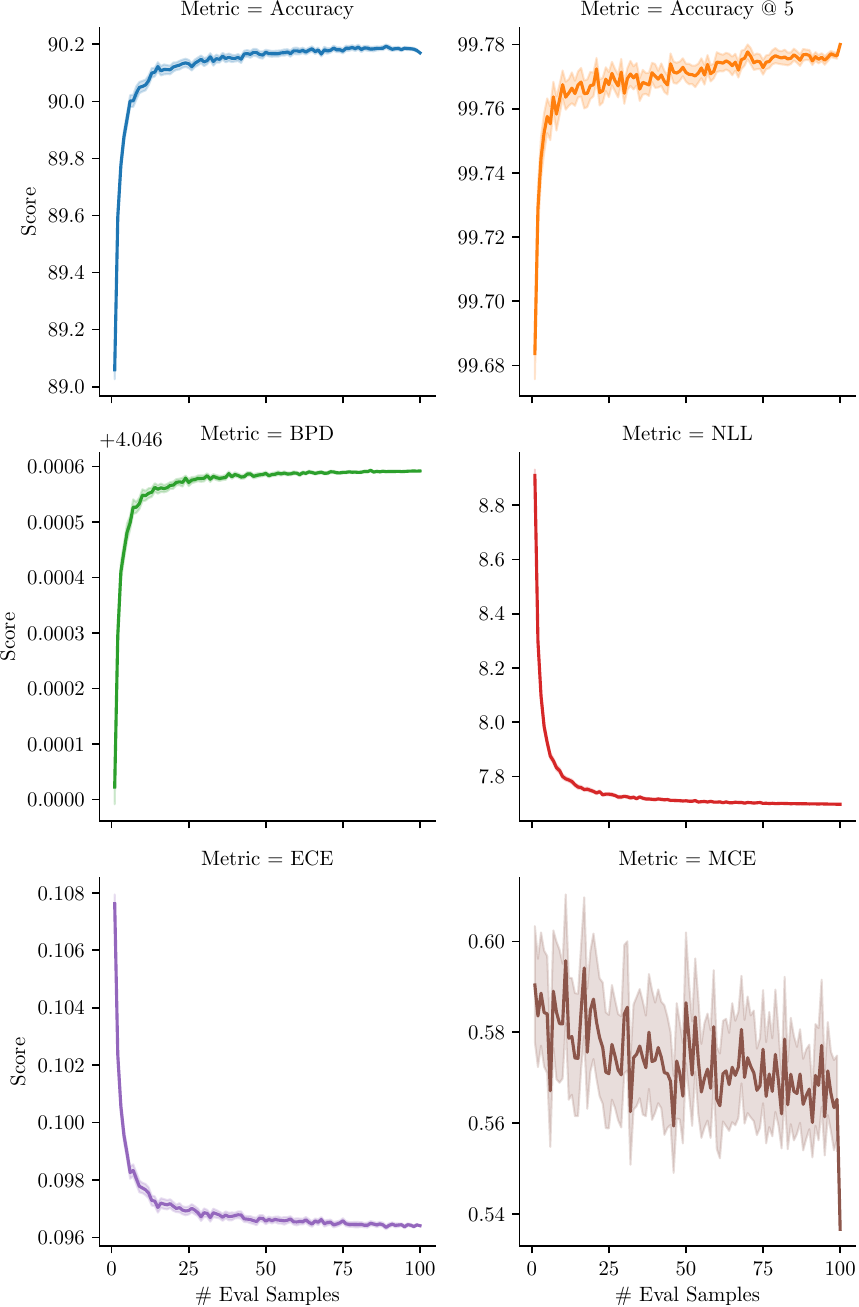}
    \caption{%
      A measure of the impact of the number of Monte Carlo samples of the DenseFlow forward pass on various performance metrics.
      The measures are used to evaluate \Ours{} trained on CIFAR-10.
    }
    \label{fig:monte-carlo-ablation}
\end{figure}

\section{Model Analysis}

\subsection{Impact of DenseFlow Monte Carlo Sampling}

On CIFAR-10, we measure the influence of the number of Monte Carlo samples on the experimental metrics measuring prediction accuracy, generation quality, and calibration quality.
As shown in \cref{fig:monte-carlo-ablation}, all scores begin to plateau around 25 samples before fully stabilizing around 100 samples.
Beyond this point, there are diminishing returns.
The number of Monte Carlo samples should be adjusted to the constraints of the application in practice---higher-stakes decision-making where calibration quality matters should allot for sampling, for instance.

\subsection{Impact of $K$-Means Initialization}

\begin{figure}
  \centering
  \includegraphics[width=0.7\linewidth]{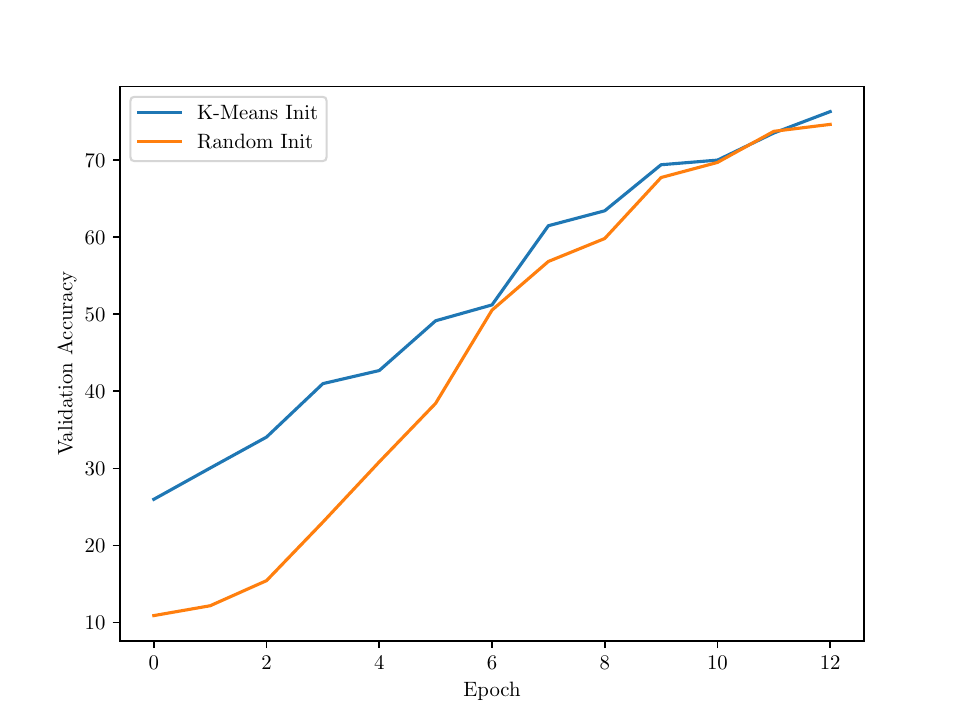}
  \caption{The impact of $K$-means initialization on \Ours{} trained on CIFAR-10.}
  \label{fig:kmeansimpact}
\end{figure}

We explore using $K$-means initialization instead of random for the prototypical Gaussian means.
As illustrated in \cref{fig:kmeansimpact}, a $K$-means initialization can jump-start training, although the model converges to the same accuracy.
This trend holds for all datasets considered and extended training.

\subsection{Loss Ablation}

We evaluate the impact of each loss function across various metrics on CIFAR-10 with results shown in \cref{tab:loss_ablate}.
The ELBO loss $\loss_\elbo$ ($p_{\Xdomain}(\data \mid y = c)$ for target class $c$) is namely responsible for reducing the BPD.
Further exploration of the losses on additional datasets is needed and we are currently compute-limited in understanding their impact more deeply.

\begin{table}[t]
  \addtolength{\tabcolsep}{4pt}
  \centering
  \begin{tabular}{@{}cccccccccc@{}}
    \toprule
    $\loss_\elbo$  & $\loss_\con$ & $\loss_\div$ & \textsc{a{\scriptsize cc}} $\uparrow$ & \textsc{bpd} $\downarrow$ \\
    \midrule
    \xmark         & \xmark       & \xmark       & 69.72                                 & 34.17                     \\
    \xmark         & \xmark       & \cmark       & 65.78                                 & 24.13                     \\
    \xmark         & \cmark       & \xmark       & 63.38                                 & 12.51                     \\
    \xmark         & \cmark       & \cmark       & 63.16                                 & 12.99                     \\
    \cmark         & \xmark       & \xmark       & 61.70                                 & 6.618                     \\
    \cmark         & \xmark       & \cmark       & 64.48                                 & 9.149                     \\
    \cmark         & \cmark       & \xmark       & 61.80                                 & 7.813                     \\
    \cmark         & \cmark       & \cmark       & 59.66                                 & 8.617                     \\
    \bottomrule                                                                                                      \\
  \end{tabular}
  \caption{%
    The impact of auxiliary loss terms when training \Ours{} on CIFAR-10.
    Due to computational constraints, each model is only trained for 11 epochs instead of fully until convergence, leading to lower accuracy scores.
    No test-time augmentation nor Monte Carlo sampling has been applied.
    All results reported in the main paper involve the $\loss_\elbo$ term, which we ablate here.
  }
  \label{tab:loss_ablate}
  \addtolength{\tabcolsep}{-4pt}
\end{table}

\subsection{Impact of Truncated Sampling of Prototypical Distributions}
Originally proposed by Brock \etal~\cite{brock2018large} as a way to improve the quality of GAN generated images, the \emph{``truncation trick''} uses a truncated version of the latent distribution $\platent$ for generation to mitigate the risk of drawing low-likelihood samples.
We analyze the results of sampling from our prototypical distributions using this \emph{``truncation trick''} across a range of truncation thresholds.
We observe that sampling from loosely truncated distributions (\ie those with a larger radius set about their mean) tends to lead to noisier images with increased visual artifacting when compared to more tightly truncated distributions.
These findings are in line Brock \etal's original assessment~\cite{brock2018large}, which notes that tighter truncation (\ie smaller radii around the mean) can improve the generated image quality at the cost of reducing variance.

\begin{figure}
  \centering
  \includegraphics[width=\linewidth,trim={0 0.75cm 0 0.2},clip]{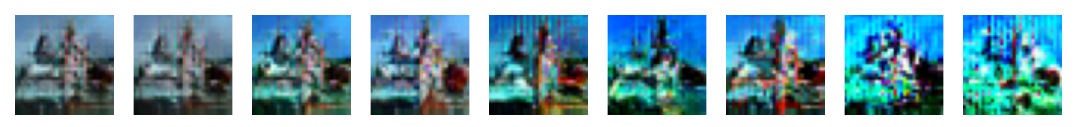}
  \\
  \centering
  \includegraphics[width=\linewidth,trim={0 0.75cm 0 0.2},clip]{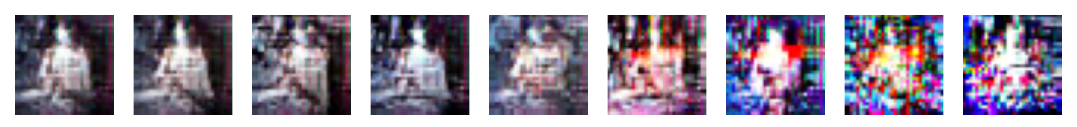}
  \\
  \centering
  \includegraphics[width=\linewidth,trim={0 0.75cm 0 0.2},clip]{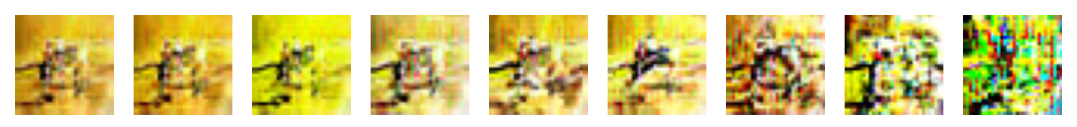}
  \\
  \centering
  \includegraphics[width=\linewidth,trim={0 0 0 0.2},clip]{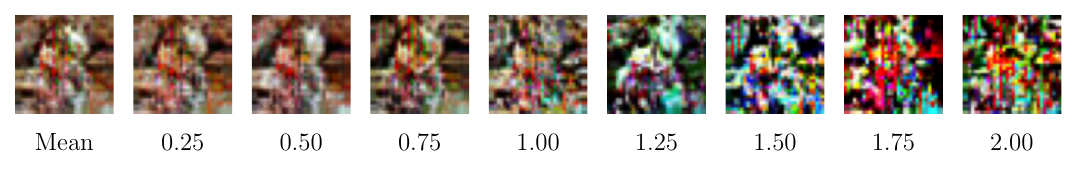}
  \caption{%
    Truncation for CIFAR-100.
    The classes represented from top to bottom are \texttt{mountain}, \texttt{woman}, \texttt{bus}, and \texttt{dinosaur}.
  }
  \label{fig:CIFAR-100_truncs}
\end{figure}

\begin{figure}
  \centering
  \includegraphics[width=\linewidth,trim={0 0.75cm 0 0.2},clip]{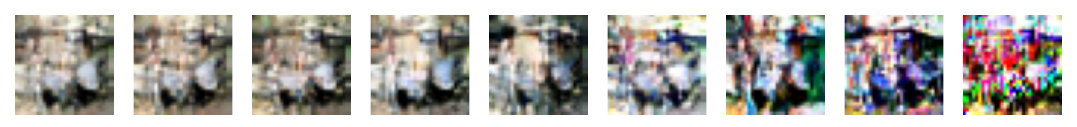}
  \\
  \centering
  \includegraphics[width=\linewidth,trim={0 0.75cm 0 0.2},clip]{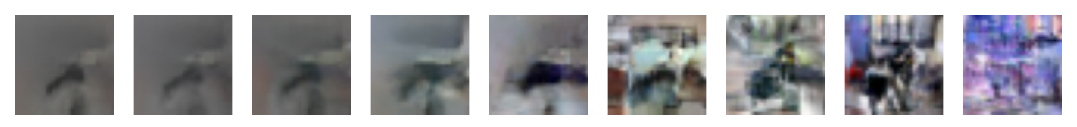}
  \\
  \centering
  \includegraphics[width=\linewidth,trim={0 0.75cm 0 0.2},clip]{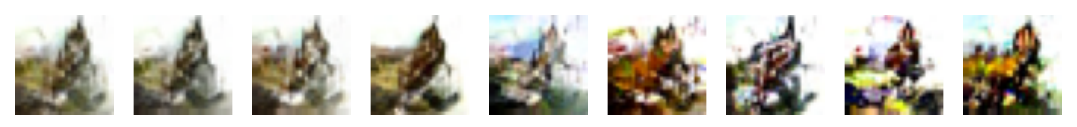}
  \\
  \centering
  \includegraphics[width=\linewidth,trim={0 0 0 0.2},clip]{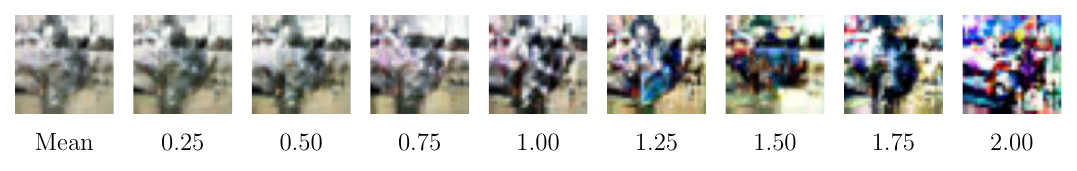}
  \caption{%
    Truncations for CIFAR-10 with consistency loss.
    The classes represented from top to bottom are \texttt{horse}, \texttt{airplane}, \texttt{bird}, and \texttt{automobile}.
  }
  \label{fig:CIFAR-10_with_const_truncs}
\end{figure}

\begin{figure}
  \centering
  \includegraphics[width=\linewidth,trim={0 0.75cm 0 0.2},clip]{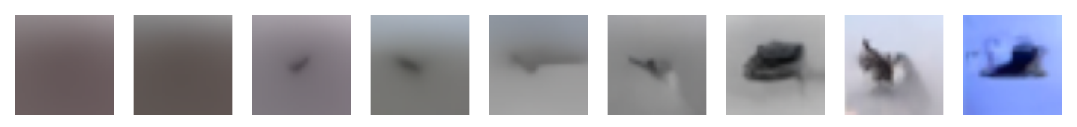}
  \\
  \centering
  \includegraphics[width=\linewidth,trim={0 0.75cm 0 0.2},clip]{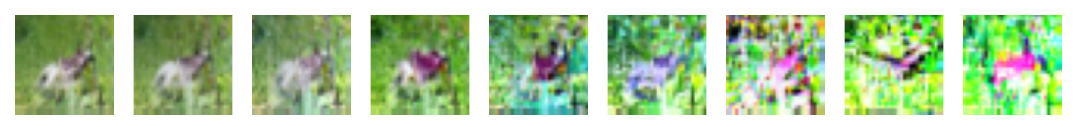}
  \\
  \centering
  \includegraphics[width=\linewidth,trim={0 0.75cm 0 0.2},clip]{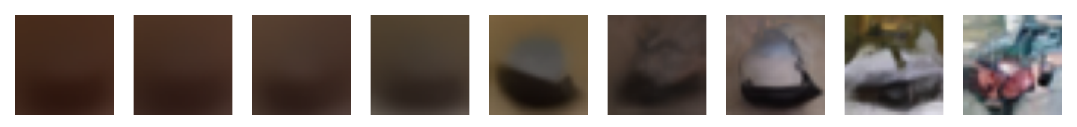}
  \\
  \centering
  \includegraphics[width=\linewidth,trim={0 0 0 0.2},clip]{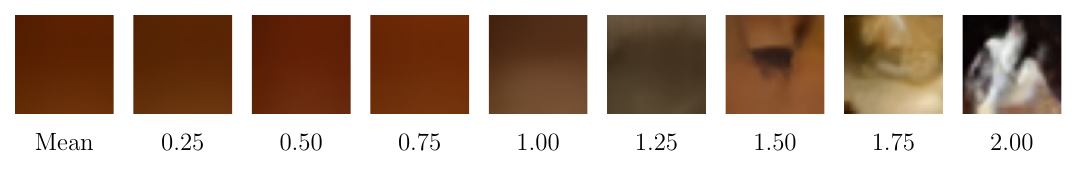}
  \caption{%
    Truncations for CIFAR-10 without consistency loss.
    The classes represented from top to bottom are \texttt{airplane}, \texttt{deer}, \texttt{automobile}, and \texttt{dog}.
  }
  \label{fig:CIFAR-10_without_const_truncs}
\end{figure}

\begin{figure}
  \centering
  \includegraphics[width=\linewidth,trim={0 0.75cm 0 0.2},clip]{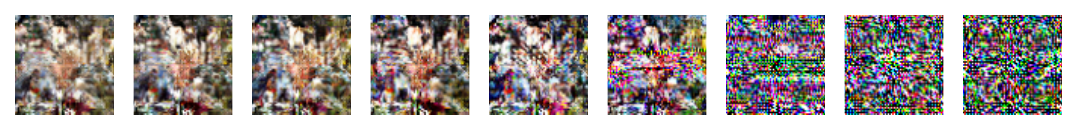}
  \\
  \centering
  \includegraphics[width=\linewidth,trim={0 0.75cm 0 0.2},clip]{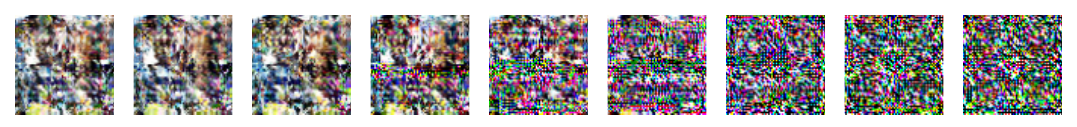}
  \\
  \centering
  \includegraphics[width=\linewidth,trim={0 0.75cm 0 0.2},clip]{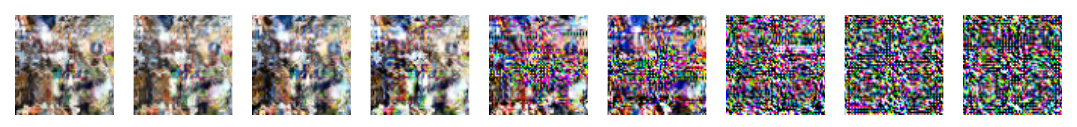}
  \\
  \centering
  \includegraphics[width=\linewidth,trim={0 0 0 0.2},clip]{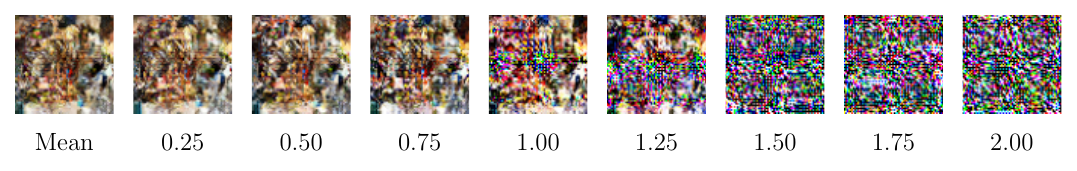}
  \caption{%
    Truncations for Oxford Pets.
    The classes represented from top to bottom are \texttt{Sphynx}, \texttt{German Shorthaired}, \texttt{Siamese}, and \texttt{Yorkshire Terrier}.
  }
  \label{fig:pets_truncs}
\end{figure}

While truncation seems to consistently have an impact across all of our experiments, the magnitude of this impact varies significantly based on the dataset, training paradigm, and even the individual components of a Gaussian mixture.
To allow comparison between truncated samples taken from different distributions with dissimilar variances, truncation values are normalized by the corresponding standard deviation for each distribution.
To see the impact of dataset choice, compare samples from \Ours{} trained on CIFAR-100 (\cf \cref{fig:CIFAR-100_truncs}) against samples from the same model trained on Oxford-IIIT Pet (\cf \cref{fig:pets_truncs}).
Sample noise scales much more aggressively with truncation in the latter than in the former, requiring a tighter truncation radius in Oxford-IIIT Pet to match the noise level of looser truncation radii on CIFAR-100.
For insight on how training plays a role, compare samples from CIFAR-10 \Ours{} trained \emph{with} (\cf \cref{fig:CIFAR-10_with_const_truncs}) and \emph{without} (\cf \cref{fig:CIFAR-10_without_const_truncs}) consistency loss.
At any given truncation level, samples from the model trained \emph{with} $\loss_\con$ are more distorted than corresponding samples from the model trained \emph{without} this term in the loss function.

Interestingly, while the discovery of differences in the significance of truncation indicates that it may be difficult to construct high quality visualizations using the truncation trick, it also suggests another possible use of truncated sampling: the assessment of the informative power of neighborhoods within the latent space.
This alternative use of truncation can be understood as complimentary to its original use; rather than attempting to find some truncation threshold that produces convincing images, instead some neighborhood is defined by a set truncation level and samples from within the neighborhood are used to determine if said neighborhood is informative.

\begin{figure}
  \centering
  \includegraphics[width=.5\linewidth]{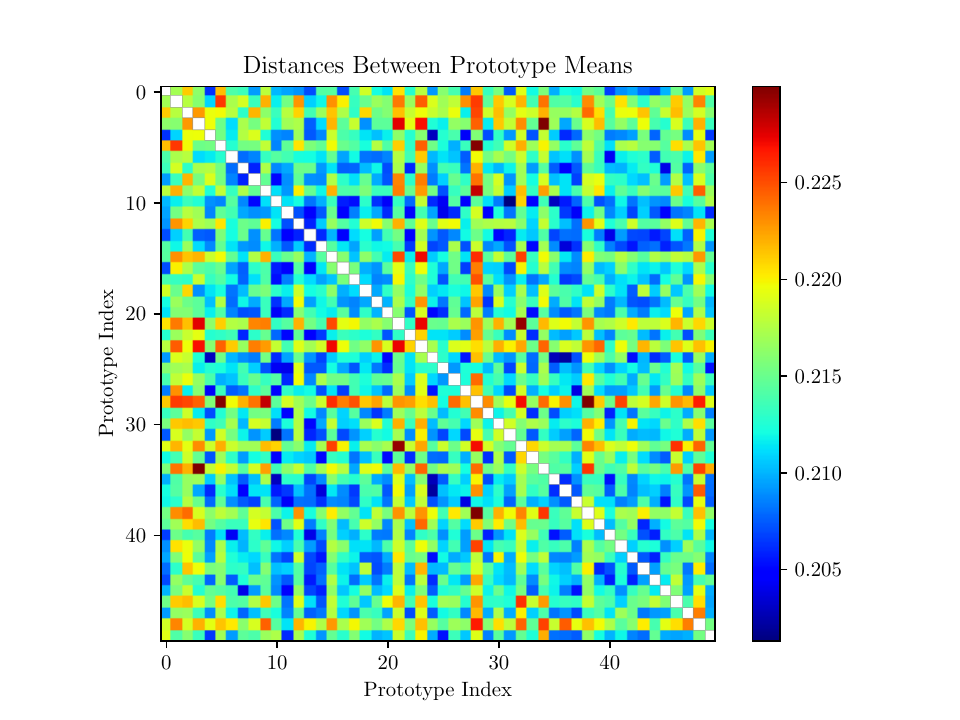}%
  \includegraphics[width=.5\linewidth]{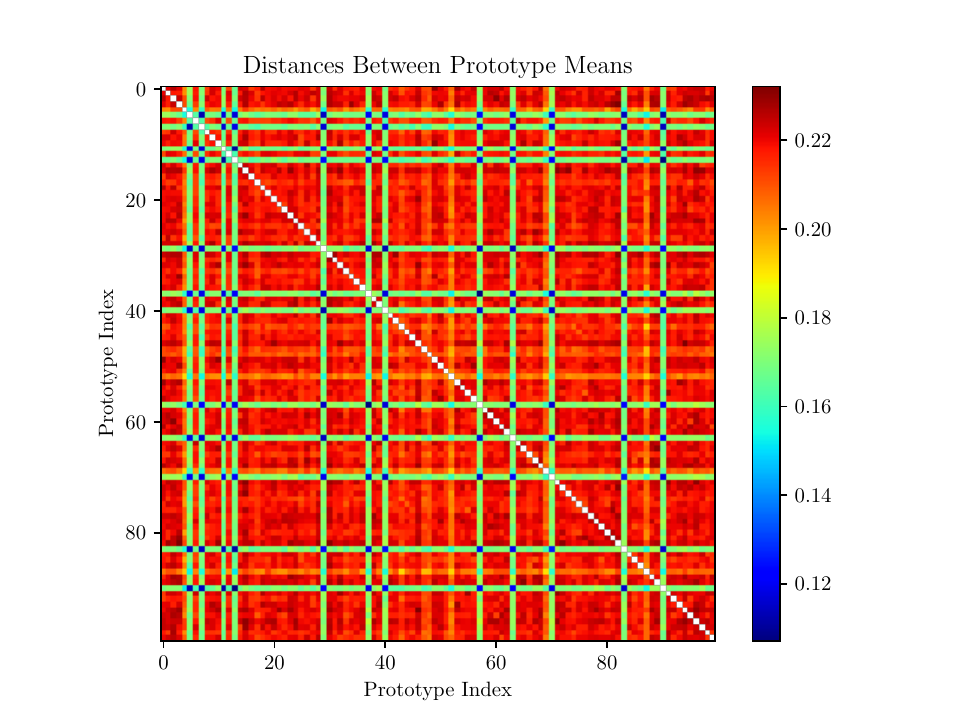}
  \caption{%
    Heatmaps showing the modified Hellinger distances between prototypes.
    On the left, \Ours{} was trained on MNIST to learn five prototypes per class.
    On the right, \Ours{} was trained on CIFAR-10 to learn ten prototypes per class.
  }
  \label{fig:hell}
\end{figure}

\subsection{Prototype Spread}

\begin{figure}
  \centering
  \includegraphics[clip, trim=0cm 0cm 0cm 1.25cm, width=.4\linewidth]{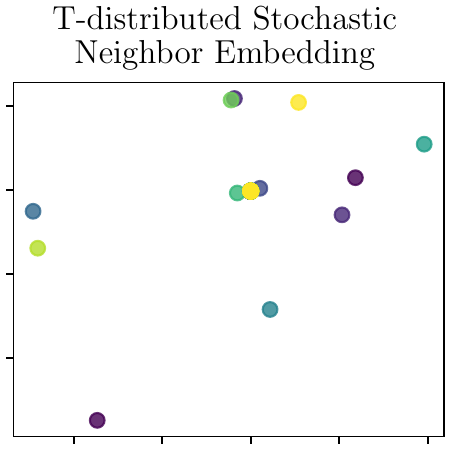}
  \caption{%
    T-SNE plot of mean points for all prototypical distribution learned by \Ours{} on CIFAR-10.
    Ten prototype distributions were learned per class.
  }
  \label{fig:tsne-cifar10}
\end{figure}

We visualize the pairwise modified Hellinger distances between all prototypical distributions learned by \Ours{} on the MNIST and CIFAR-10 datasets with the respective heatmaps in \cref{fig:hell}.
The diagonals are masked for simplicity since they are uninformative.
We see a spread in similarities among prototypical distributions, which is corroborated by the T-SNE plot in \cref{fig:tsne-cifar10}.

\subsection{Impact of Consistency Loss}
Continuing our discussion from the main paper on the impact of the consistency loss term $\loss_\con$, we quickly recap that prototypes learned \emph{with} $\loss_\con$ have somewhat good-looking mean-point visualizations, but sampling na\"ively from those distributions leads to unacceptably noisy images, motivating our use and discussion of truncated sampling when training with $\loss_\con$.
This is contrasted by the uninformative, blurry prototype mean images learned \emph{without} $\loss_\con$ which, nonetheless, yield visually and semantically meaningful samples.

Since this behavior is so consistent across prototypes within each model, we hypothesized that the consistency term affected the covariances of the learned prototypes.
However, there was no significant difference in the covariances' $L^2$ norms, and the curse of dimensionality makes it exceedingly unlikely any existing difference would be statistically significant.

\begin{figure}
  \centering
  \includegraphics[width=.75\linewidth]{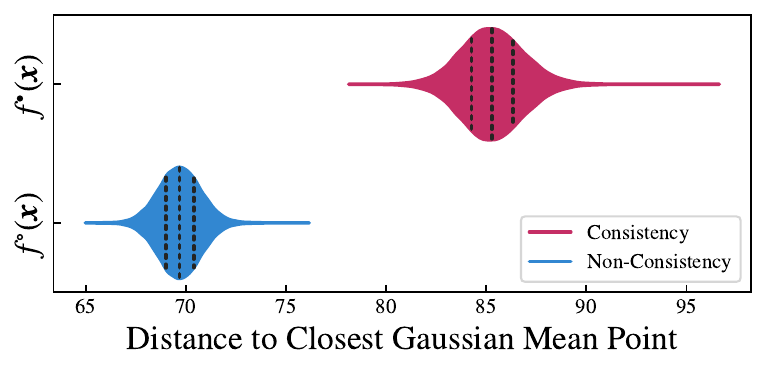}
  \caption{%
    Distances from each embedded point to its nearest prototypical mean point plotted for a model trained \emph{with} (in red) and \emph{without} (in blue) consistency regularization.
    Quartiles are shown as dashed lines.
  }
  \label{fig:distance_closest_proto}
\end{figure}

Instead, we realized that the presence of $\loss_\con$ affected the \emph{spacial arrangement} of the embedded latent points $f(\data) \in \latent$ relative to the learned prototype distributions' locations in $\latent$.
For each data point $\data_i \in \domain$, we compute $\delta_i = \min_{\proto_{c, k}}\norm{\data_i - \mean_{c, k}}$ and plot these distributions in \cref{fig:distance_closest_proto}.
The model $\wocon{f}$ trained \emph{without} $\loss_\con$ tended to embed the data points significantly closer in $\latent$ to the mean of a prototypical distribution than $\wcon{f}$ did.
This could help explain why samples from $\wocon{f}$ seem to look more \emph{``real''} while samples from $\wcon{f}$ look \emph{``noisy.''}

This further works to point out the flaw with point-based prototypical explainers.
The fact that samples from $\wcon{\proto_{c, k}}$ look so bad suggests that the regions of $\latent$ those distributions preside over are not informative---at least to a human---about the task being performed.
However, this would never have been discovered if only information about the \emph{mean points} of these prototypes was known, as they seem substantially less noisy.
Similarly, despite the uninformative mean points learned by removing $\loss_\con$, the actual samples from $\wocon{\proto_{c, k}}$ without consistency are consistently and undeniably more meaningful.

\subsection{Pruning Intuition}
\begin{figure}
  \centering
  \includegraphics[width=.75\linewidth]{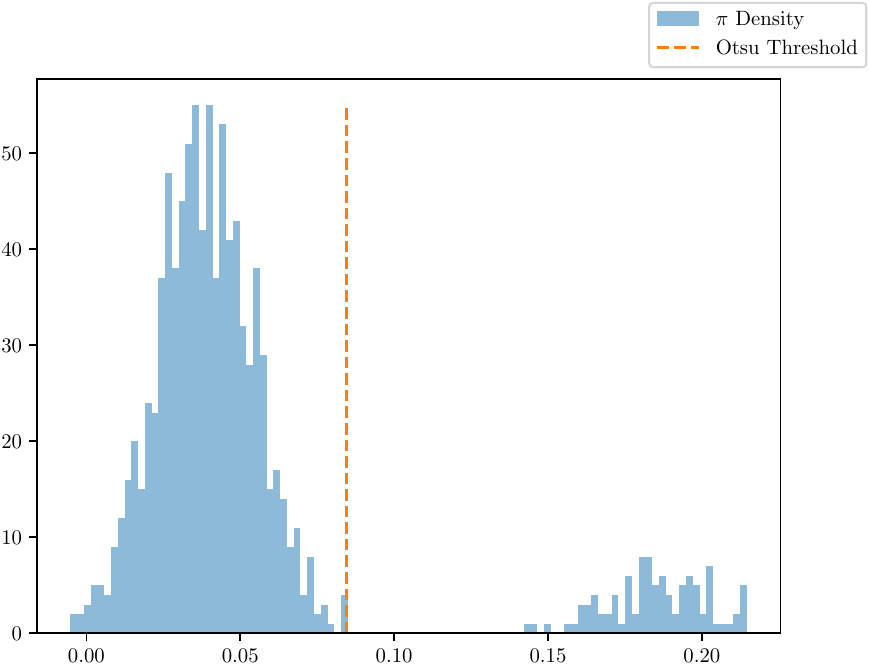}
  \caption{%
    Histogram of learned Gaussian weights $\pi_{c,k}$ for \Ours{} trained on CIFAR-100.
    The majority of prototypes are elected to be pruned using Otsu's method.
  }
  \label{fig:cifar100-otsu-pruning}
\end{figure}

We consider pruning the learned prototypical distributions to reduce the number of elements required for interpretation and the number of parameters in the model.
Recall that we do so by removing prototypes that have a weight $\pi_{c,k}$ less than some $\varepsilon$.
Otsu's method is a good candidates to determine this threshold as such weight distributions are typically bimodal, as shown in \cref{fig:cifar100-otsu-pruning}.
This approach to finding a threshold minimizes the sample variance of the points on either side of the threshold (intra-class intensity variance)---it discovers the point that indicates whether a prototype is important for prediction.

%% file: tables/pp_results.tex
\begin{table}[b]
    \ra{1.2}
    \centering
    \addtolength{\tabcolsep}{2pt}
    \begin{tabular}{@{}lcccc@{}}
        Model & Resolution & Visualization & \textsc{Acc} $\uparrow$ & \textsc{Acc} @ 5 $\uparrow$ \\
        \toprule
          \Ours{} & {$64\times 64$} & Flow Inverse & 50.58 & 77.46 \\
          ProtoPNet & {$224\times 224$} & Proto Projection & 76.42 & 91.78 \\
        \bottomrule \\
    \end{tabular}
    \caption{Comparison on CUB-200-2011 with 10 prototypes learned per class.}
    \label{tab:ppResults}
    \addtolength{\tabcolsep}{-2pt}
\end{table}

%% file: tables/clf_results_full.tex
\begin{table}[t]
    \ra{1.2}
    \tiny
    \centering
    \addtolength{\tabcolsep}{2pt}
    \begin{tabular}{@{}lllcccccc@{}}
        \toprule
        Dataset & Res & Model & Proto-Based & Flow-Based & Acc $\uparrow$ & BPD $\downarrow$ & ECE $\downarrow$ & MCE $\downarrow$ \\
        \midrule
        \multirow{5}{*}{\rotatebox{0}{MNIST}} & \multirow{5}{*}{$28\times 28$} & 
            \Ours{} (Ours) & \cmark & \cmark & {99.36} & \textbf{0.535} & 0.006 & 0.587 \\
          && FlowGMM~\cite{izmailov2020semi} & \xmark & \cmark & \textbf{99.63} & --- & \textbf{0.004}* & --- \\
          && Fetaya et al.~\cite{DBLP:conf/iclr/FetayaJGZ20} & \xmark & \cmark & 99.30 & {1.00} & --- & --- \\
          && SCNF-GLOW~\cite{atanov2019semi} & \xmark & \cmark & 88.44 & 1.15 & --- & --- \\
          && SCNF-GMM~\cite{atanov2019semi} & \xmark & \cmark & 83.10 & 1.14 & --- & --- \\[1.5ex]
        \multirow{11}{*}{\rotatebox{0}{CIFAR-10}} & \multirow{11}{*}{$32\times 32$} &
            \Ours{} (Ours) & \cmark & \cmark & \textbf{91.54} & 3.95 & 0.083 & \textbf{0.494} \\
          && IB-INN ($\gamma \rightarrow \infty$)~\cite{NEURIPS2020_593906af} & \xmark & \cmark & 91.28 & 17.3 & 0.81 & 13.9 \\
          && IB-INN ($\gamma=1$)~\cite{NEURIPS2020_593906af} & \xmark & \cmark & 89.73 & 5.25 & 0.54 & 3.25 \\
          && FlowGMM~\cite{izmailov2020semi} & \xmark & \cmark & 88.44 & --- & \textbf{0.038}* & --- \\
          && Fetaya et al.~\cite{DBLP:conf/iclr/FetayaJGZ20} & \xmark & \cmark & 84.00 & \textbf{3.53} & --- & --- \\[1ex]
          && KMEx~\cite{gautam2023prototypical} & \cmark & \xmark & 85.3 & --- & --- & --- \\
          && ProtoPNet~\cite{protopnet,gautam2023prototypical} & \cmark & \xmark & 84.9 & --- & --- & --- \\
          && FLINT~\cite{parekh2021framework} & \cmark & \xmark & 84.7 & --- & --- & --- \\
          && ProtoVAE~\cite{gautam2022protovae} & \cmark & \xmark & 84.6 & --- & --- & --- \\[1ex]
          && ViT-H/14~\cite{DBLP:conf/iclr/DosovitskiyB0WZ21} & \xmark & \xmark & \textit{{99.50}} & --- & --- & --- \\[1.5ex]
        \multirow{4}{*}{\rotatebox{0}{CIFAR-100}} & \multirow{4}{*}{$32\times 32$} &
            \Ours{} (Ours) & \cmark & \cmark & \textbf{69.80} & 5.03 & \textbf{0.292} & \textbf{0.637} \\
          && IB-INN ($\gamma \rightarrow \infty$)~\cite{NEURIPS2020_593906af} & \xmark & \cmark & 66.22 & 18.4 & 0.62 & 16.8 \\
          && IB-INN ($\gamma=1$)~\cite{NEURIPS2020_593906af} & \xmark & \cmark & 57.43 & \textbf{4.93} & 0.58 & 7.04 \\[1ex]
          && EffNet-L2 (SAM)~\cite{DBLP:conf/iclr/ForetKMN21} & \xmark & \xmark & \textit{{96.08}} & --- & --- & --- \\[1.5ex]
        \multirow{2}{*}{Flowers-102} & \multirow{2}{*}{$64\times 64$} & \Ours{} (Ours) & \cmark & \cmark & {59.80} & 13.5 & {0.141} & {0.295} \\
        && CCT-14/7×2~\cite{hassani2021escaping} & \xmark & \xmark & \textit{{99.76}}** & --- & --- & --- \\[1.5ex]
        \multirow{2}{*}{Oxford-IIIT Pet} & \multirow{2}{*}{$64\times 64$} & \Ours{} (Ours) & \cmark & \cmark & {53.58} & 4.89 & {0.459} & {0.736} \\
        && OmniVec~\cite{srivastava2024omnivec} & \xmark & \xmark & \textit{{99.2}}** & --- & --- & --- \\
        \bottomrule \\
    \end{tabular}
    \caption{Results of normalizing-flow-based joint generative and predictive models across image classification tasks with non-generative classifier baselines. Results reported from other prototypical neural networks are shown for CIFAR-10. \Ours{} achieves state-of-the-art accuracy while retaining highly competitive density estimation and calibration scores.
    *The result is reported with temperature-based scaling of GMM variances~\cite{izmailov2020semi}. **A larger resolution of each image is used for both training and testing.%
    }
    \label{tab:clfResultsFull}
    \addtolength{\tabcolsep}{-2pt}
\end{table}